\documentclass[journal,comsoc]{IEEEtran}
\usepackage[T1]{fontenc}

\usepackage{cite}

\usepackage{color}

\ifCLASSINFOpdf
  \usepackage[pdftex]{graphicx}
\else
  \usepackage[dvips]{graphicx}
\fi
\usepackage{amsmath}
\usepackage[cmintegrals]{newtxmath}
\usepackage{booktabs} 
\usepackage{multirow}
\usepackage{supertabular}

\usepackage{makecell}

\usepackage[linesnumbered,ruled]{algorithm2e}
\SetKwProg{Fn}{Function}{}{end}\SetKwFunction{FRecurs}{FnRecursive}%

\usepackage{flushend}

\ifCLASSOPTIONcompsoc
  \usepackage[caption=false,font=normalsize,labelfont=sf,textfont=sf]{subfig}
\else
  \usepackage[caption=false,font=footnotesize]{subfig}
\fi
\begin{document}
%
\title{Push and Pull Search for Solving Constrained Multi-objective Optimization Problems}
%
%
%

\author{Zhun Fan,~\IEEEmembership{Senior Member,~IEEE,}
        Wenji Li,
        Xinye Cai,
        Hui Li,
        Caimin Wei, 
        Qingfu Zhang,~\IEEEmembership{Fellow,~IEEE,} 
        Kalyanmoy Deb,~\IEEEmembership{Fellow,~IEEE,} 
        and Erik D. Goodman
\thanks{}
}

%
%

\markboth{Journal of \LaTeX\ Class Files,~Vol.~14, No.~8, August~2015}%
{Shell \MakeLowercase{\textit{et al.}}: Bare Demo of IEEEtran.cls for IEEE Communications Society Journals}
%



\maketitle

\begin{abstract}
This paper proposes a push and pull search (PPS) framework for solving constrained multi-objective optimization problems (CMOPs). To be more specific, the proposed PPS divides the search process into two different stages, including the push and pull search stages. In the push stage, a multi-objective evolutionary algorithm (MOEA) is adopted to explore the search space without considering any constraints, which can help to get across infeasible regions very fast and approach the unconstrained Pareto front. Furthermore, the landscape of CMOPs with constraints can be probed and estimated in the push stage, which can be utilized to conduct the parameters setting for constraint-handling approaches applied in the pull stage. Then, a constrained multi-objective evolutionary algorithm (CMOEA) equipped with an improved epsilon constraint-handling is applied to pull the infeasible individuals achieved in the push stage to the feasible and non-dominated regions. Compared with other CMOEAs, the proposed PPS method can more efficiently get across infeasible regions and converge to the feasible and non-dominated regions by applying push and pull search strategies at different stages. To evaluate the performance regarding convergence and diversity, a set of benchmark CMOPs is used to test the proposed PPS and compare with other five CMOEAs, including MOEA/D-CDP, MOEA/D-SR, C-MOEA/D, MOEA/D-Epsilon and MOEA/D-IEpsilon. The comprehensive experimental results demonstrate that the proposed PPS achieves significantly better or competitive performance than the other five CMOEAs on most of the benchmark set. 
\end{abstract}

\begin{IEEEkeywords}
Push and Pull Search, Constraint-handling Mechanisms, Constrained Multi-objective Evolutionary Algorithms. 
\end{IEEEkeywords}

%
\IEEEpeerreviewmaketitle

\section{Introduction}
%
%
%
%
\IEEEPARstart{M}{any} real-world optimization problems can be summarized as optimizing a number of conflicting objectives simultaneously with a set of equality and/or inequality constraints, which are denoted as constrained multi-objective optimization problems (CMOPs). Without lose of generality, a CMOP considered in this paper can be defined as follows \cite{deb2001multi}:

\begin{equation}
\label{equ:cmop_definition}
\begin{cases}
\mbox{minimize} &\mathbf{F}(\mathbf{x}) = {(f_{1}(\mathbf{x}),\ldots,f_{m}(\mathbf{x}))} ^ {T} \\
\mbox{subject to} & g_i(\mathbf{x}) \ge 0, i = 1,\ldots,q \\
& h_j(\mathbf{x}) = 0, j= 1,\ldots,p \\
&\mathbf{x} \in {\mathbb{R}^n}
\end{cases}
\end{equation}
where $F(\mathbf{x}) = ({f_1}(\mathbf{x}),{f_2}(\mathbf{x}), \ldots ,{f_m}(\mathbf{x})) ^ T$ is an $m$-dimensional objective vector, and $F(\mathbf{x})\in \mathbb{R} ^m$. ${g_i}(\mathbf{x}) \ge 0$ is an inequality constraint, and $q$ is the number of inequality constraints. ${h_j}(\mathbf{x})=0$ is an equality constraint, and $p$ represents the number of equality constraints. $\mathbf{x} \in \mathbb{R}^n$ is an $n$-dimensional decision vector.

When solving CMOPs with both inequality and equality constraints, we usually convert the equality constraints into inequality ones by introducing an extremely small positive number $\epsilon$. The detailed transformation is given as follows:

\begin{eqnarray}
\label{equ:inequal2equal}
h_j(\mathbf{x})' \equiv \epsilon - |h_j(\mathbf{x})|  \ge 0
\end{eqnarray}

To deal with a set of constraints in CMOPs, the overall constraint violation is a widely used approach, which summarizes them into a single scalar, as follows: 
\begin{eqnarray}
\label{equ:constraint}
\phi(\mathbf{x}) = \sum_{i=1}^{q} |\min(g_i(\mathbf{x}),0)| + \sum_{j = 1}^{p} |\min(h_j(\mathbf{x})',0)|
\end{eqnarray}

Given a solution $\mathbf{x}^{k} \in {\mathbb{R}^n}$, if $\phi(\mathbf{x}^{k}) = 0$, $\mathbf{x}^{k}$ is a feasible solution. All the feasible solutions constitute a feasible region $S$, which is defined as $S = \{\mathbf{x} | \phi(\mathbf{x}) = 0, \mathbf{x \in {\mathbb{R}^n} \}}$. For any two solutions $\mathbf{x}^a, \mathbf{x}^b \in S$, $\mathbf{x}^a$ is said to dominate $\mathbf{x}^b$ if $f_i(\mathbf{x}^a) \le f_i(\mathbf{x}^b)$ for each $i \in \{1,...,m\}$ and $F(\mathbf{x}^a) \neq F(\mathbf{x}^b)$, denoted as $\mathbf{x}^a \preceq \mathbf{x}^b$. If there is no other solution in $S$ dominating solution $\mathbf{x}^*$, then $\mathbf{x}^*$ is called a Pareto optimal solution. All of the Pareto optimal solutions constitute a Pareto optimal set ($PS$). The mapping of the $PS$ in the objective space is called a Pareto optimal front ($PF$), which is defined as $PF = \{F(\mathbf{x})| \mathbf{x} \in PS\}$.


A key issue in CMOEAs is to maintain a balance the objectives and constraints. In fact, most constraint-handling mechanisms in evolutionary computation are applied for this purpose.

For example, the penalty function approach adopts a penalty factor $\lambda$ to maintain the balance between the objectives and constraints. It converts a CMOP into an unconstrained MOP by adding the overall constraint violation multiplied by a predefined penalty factor $\lambda$ to each objective \cite{CoelloCoello20021245}. In the case of $\lambda = \infty$, it is called a death penalty approach \cite{bdack1991survey}, which means that infeasible solutions are totally unacceptable. If $\lambda$ is a static value during the search process, it is called a static penalty approach \cite{homaifar1994constrained}. If $\lambda$ is changing during the search process, it is called a dynamic penalty approach \cite{joines1994use}. In the case of $\lambda$ is changing according to the information collected from the search process, it is called an adaptive penalty approach \cite{bean1993dual,coit1996adaptive,ben1997genetic,4799193}.


In oder to avoid tuning the penalty factors, another type of constraint-handling methods is suggested, which compares the objectives and constraints separately. Representative examples include constraint dominance principle (CDP) \cite{996017}, epsilon constraint-handling method (EC) \cite{takahama2005constrained}, stochastic ranking approach (SR) \cite{runarsson2000stochastic}, and so on.

In CDP \cite{996017}, three basic rules are adopted to compare any two solutions. In the first rule, given two solutions $\mathbf{x}_i, \mathbf{x}_j \in {\mathbb{R}^n}$, if $\mathbf{x}_i$ is feasible and $\mathbf{x}_j$ is infeasible, $\mathbf{x}_i$ is better than $\mathbf{x}_j$. In the case of $\mathbf{x}_i$ and $\mathbf{x}_j$ are both infeasible, the one with a smaller constraint violation is better. In the last rule, $\mathbf{x}_i$ and $\mathbf{x}_j$ are both feasible, and the one dominating the other is better. CDP is a popular constraint-handling method, as it is simple and has no extra parameters. However, it is not suitable for solving CMOPs with very small and narrow feasible regions \cite{Fan2016Angle}. For many generations, most or even all solutions in the working population are infeasible when solving CMOPs with this property. In addition, the diversity of the working population can hardly be well maintained, because the selection of solutions is only based on the constraint violations according to the second rule of CDP.

In order to solve CMOPs with small and narrow feasible regions, the epsilon constraint-handling (EC) \cite{takahama2005constrained} approach is suggested. It is similar to CDP except for the relaxation of the constraints. In EC, the relaxation of the constraints is controlled by the epsilon level $\varepsilon$, which can help to maintain the diversity of the working population in the case when most solutions are infeasible. To be more specific, if the overall constraint violation of a solution is less than $\varepsilon$, this solution is deemed as feasible. The epsilon level $\varepsilon$ is a critical parameter in EC. In the case of $\varepsilon = 0$, EC is the same as CDP. Although EC can be used to solve CMOPs with small feasible regions, controlling the value of $\varepsilon$ properly is non-trivial at all.

Both CDP \cite{996017} and EC \cite{takahama2005constrained} first compare the constrains, then compare the objectives. SR \cite{runarsson2000stochastic} is different from CDP and EC in terms of the order of comparison. It adopts a probability parameter $p_f \in [0,1]$ to decide if the comparison is based on objectives or constraints. For any two solutions, if a random number is less than $p_f$, the one with the non-dominated objectives is better. The comparison is based on objectives. On the contrary, if the random number is greater than $p_f$, the comparison is first based on the constraints, then on the objectives, which is the same as that of CDP. In the case of $p_f = 0$, SR is equivalent to CDP.


In recent years, many work has been done in the field of many-objective evolutionary algorithms (MaOEAs) \cite{li2015many}, which gives us a new way to solve CMOPs. In order to balance the constraints and the objectives, some researchers adopt multi-objective evolutionary algorithms (MOEAs) or MaOEAs (in the case of the number of objectives is greater than three) to deal with constraints \cite{MezuraMontes:2011cj}. For an M-objective CMOP, its constraints can be converted into one or $k$ extra objectives. Then the M-objective CMOP is transformed into an $(M + 1)$- or $(M + k)$-objective unconstrained MOP, which can be solved by MOEAs or MaOEAs. Representative examples include Cai and Wang’s Method (CW) \cite{cai2006multiobjective} and the in-feasibility driven evolutionary algorithms (IDEA) \cite{ray2009infeasibility}. 


To maintain a good balance between the constraints and the objectives, some researchers combine several constraint-handling mechanisms, which can be further divided into two categories, including adopting different constraint-handling mechanisms in different evolutionary stages or in different subproblems. For example, the adaptive trade-off model (ATM) \cite{wang2008adaptive} uses two different constraint-handling mechanisms, including multi-objective approach and the adaptive penalty functions in different evolutionary stages. The ensemble of constraint-handling methods (ECHM) \cite{qu2011constrained} uses three different constraint-handling techniques, including epsilon constraint-handling (EC) \cite{takahama2005constrained}, self-adaptive penalty functions (SP) \cite{4799193} and superiority of feasible solutions (SF) \cite{deb2000efficient}. Three sub-populations are generated in ECHM, and each sub-population uses a different constraint-handling method. 


In this paper, we propose a biphasic CMOEA, namely push and pull search (PPS), to balance the objectives and the constraints. Unlike the above-mentioned constraint-handling methods, the PPS divide the search process into two different stages. In the first stage, only the objectives are optimized, which means the working population is pushed toward the unconstrained PF without considering any constraints. Furthermore, the landscape of constraints in CMOPs can be estimated in the push stage, which can be applied to conduct the parameters setting of constraint-handling approaches applied in the pull stage. In the pull stage, an improved EC \cite{takahama2005constrained} constraint-handling approach is adopted to pull the working population to the constrained PF. In summary, it provides a new framework and has the following potential advantages.

\begin{enumerate}
  \item It has the ability to get across large infeasible regions in front of the constrained PF. Since the constraints are ignored in the push stage, the infeasible regions in front of the real PF present no barriers for the working population.

    \item It facilitates the parameters setting in the constraint-handling methods. Since the landscape of constraints has already been explored by the push process, a lots of information can be discovered and gathered to guide the search and parameter setting in the pull stage. 
\end{enumerate}

The rest of the paper is organized as follows. Section \ref{sec:pps} introduces the general idea of PPS. Section \ref{sec:proposed-method} gives an instantiation of PPS in the framework of MOEA/D, called PPS-MOEA/D. Section \ref{sec:exper} designs a set of experiments to compare the proposed PPS-MOEA/D with five other CMOEAs, including MOEA/D-IEpsilon, MOEA/D-Epsilon, MOEA/D-SR, MOEA/D-CDP and C-MOEA/D. Finally, conclusions are made in section \ref{sec:conc}. 

\section{The General Framework of Push and Pull Search}
\label{sec:pps}

\begin{figure*}
\begin{tabular}{cc}
\begin{minipage}[t]{0.28\linewidth}
\includegraphics[width = 5cm]{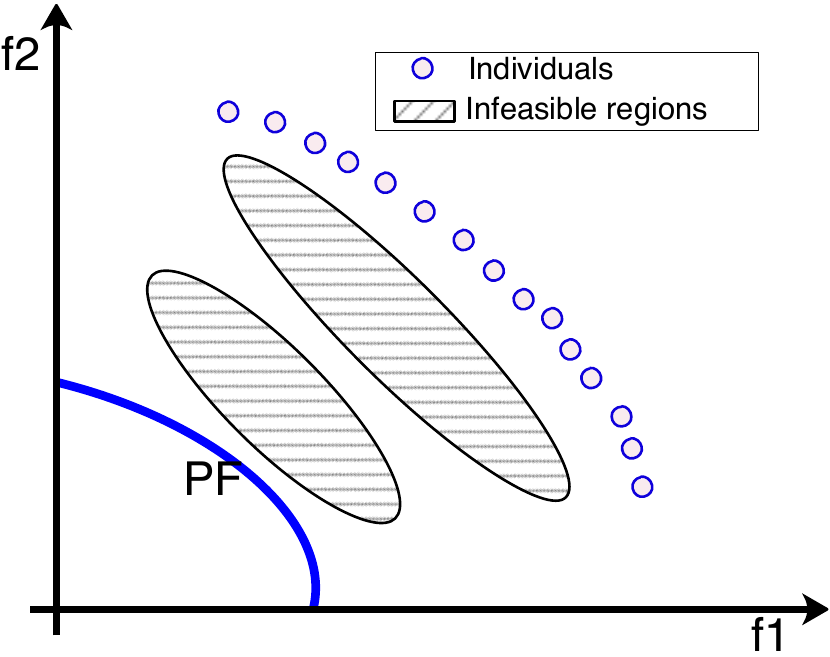}\\
\centering{\scriptsize{(a)}}
\end{minipage}
\hspace{0.5cm}
\begin{minipage}[t]{0.28\linewidth}
\includegraphics[width = 5cm]{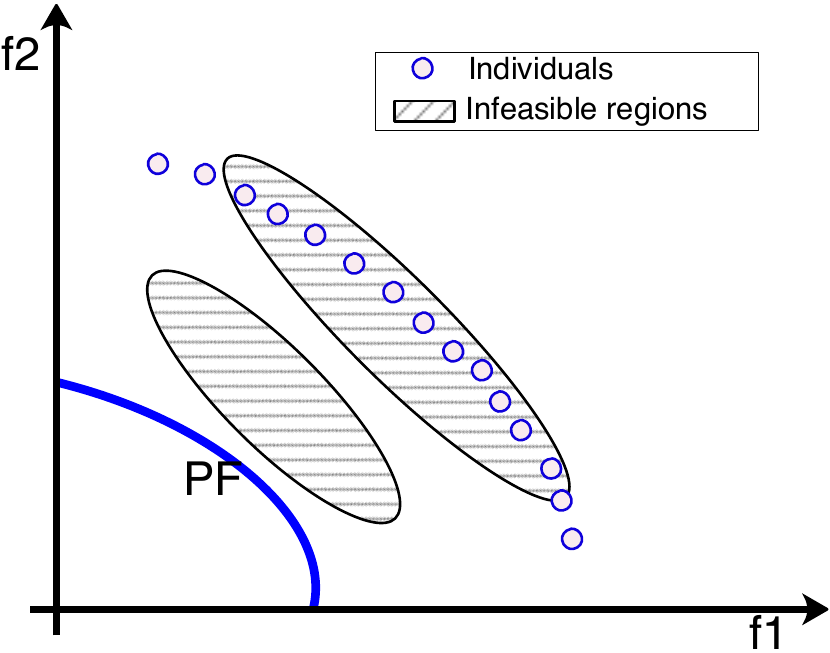}\\
\centering{\scriptsize{(b)}}
\end{minipage}
\hspace{0.5cm}
\begin{minipage}[t]{0.28\linewidth}
\includegraphics[width = 5cm]{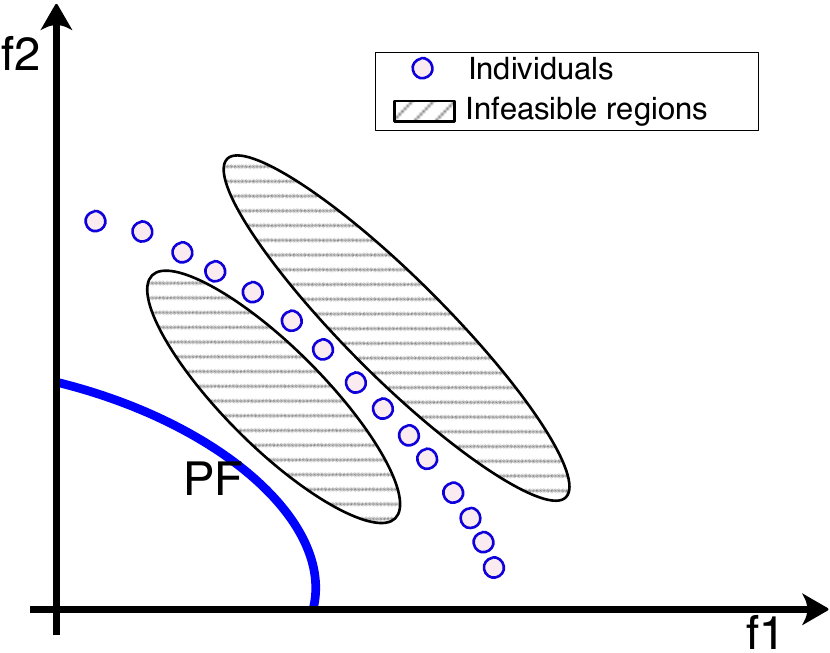}\\
\centering{\scriptsize{(c)}}
\end{minipage}
\end{tabular}

\vspace{0.5cm}
\begin{tabular}{cc}
\begin{minipage}[t]{0.28\linewidth}
\includegraphics[width = 5cm]{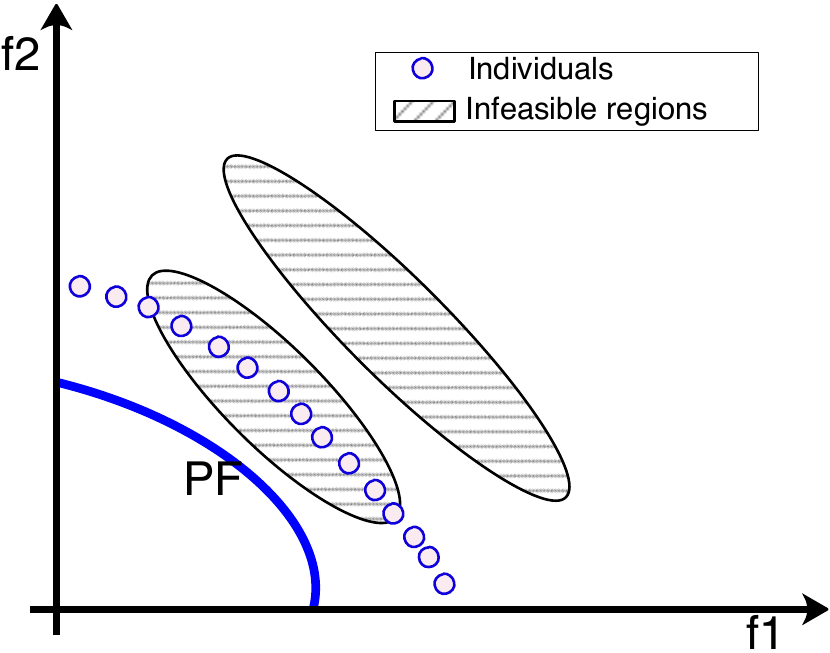}\\
\centering{\scriptsize{(d)}}
\end{minipage}
\hspace{0.5cm}
\begin{minipage}[t]{0.28\linewidth}
\includegraphics[width = 5cm]{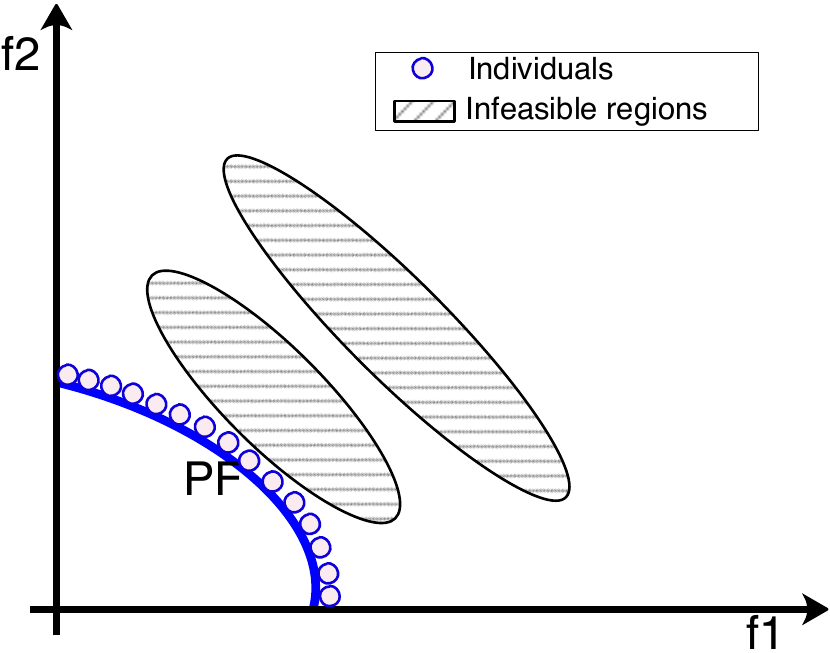}\\
\centering{\scriptsize{(e)}}
\end{minipage}
\hspace{0.5cm}
\begin{minipage}[t]{0.28\linewidth}
\includegraphics[width = 5cm]{figures/evolutionary-process/case-a-5.eps}\\
\centering{\scriptsize{(f)}}
\end{minipage}
\end{tabular}

\caption{\label{fig:case-a} Infeasible regions block the way towards the PF, and the unconstrained PF is the same as the constrained PF. (a)-(e) show the different stages of the push search process, and the working population can get across the infeasible regions without any extra efforts dealing with constraints. (f) shows the pull search process, which is the same as (e) in this particular case, since the real PF is the same as the unconstrained PF, and has already been achieved by the working population in the push search process.}
\end{figure*}

\begin{figure*}
\begin{tabular}{cc}
\begin{minipage}[t]{0.28\linewidth}
\includegraphics[width = 5cm]{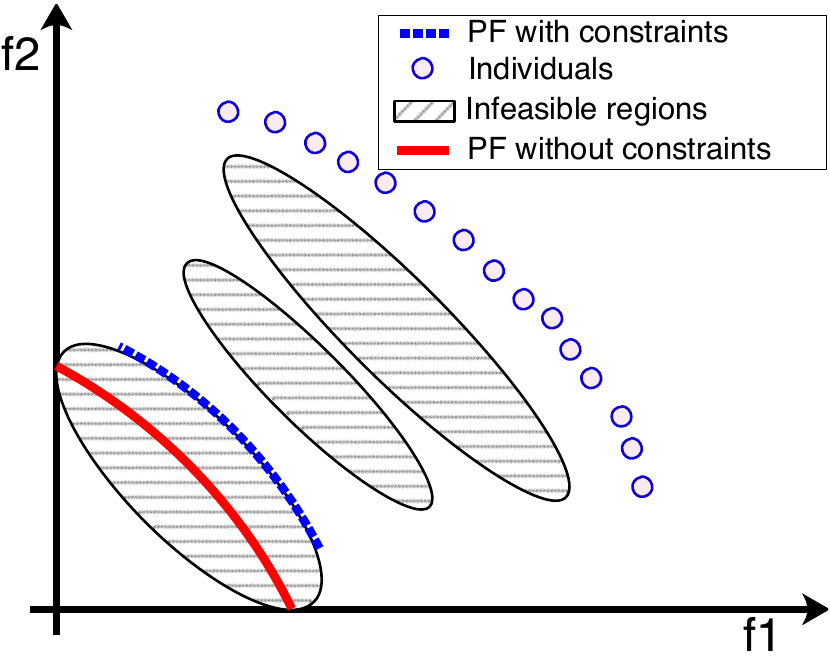}\\
\centering{\scriptsize{(a)}}
\end{minipage}
\hspace{0.5cm}
\begin{minipage}[t]{0.28\linewidth}
\includegraphics[width = 5cm]{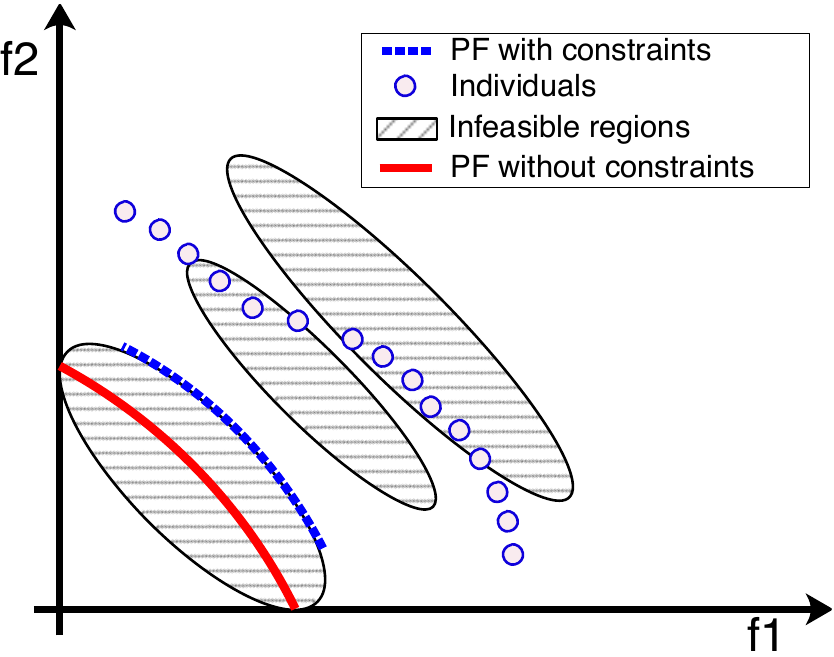}\\
\centering{\scriptsize{(b)}}
\end{minipage}
\hspace{0.5cm}
\begin{minipage}[t]{0.28\linewidth}
\includegraphics[width = 5cm]{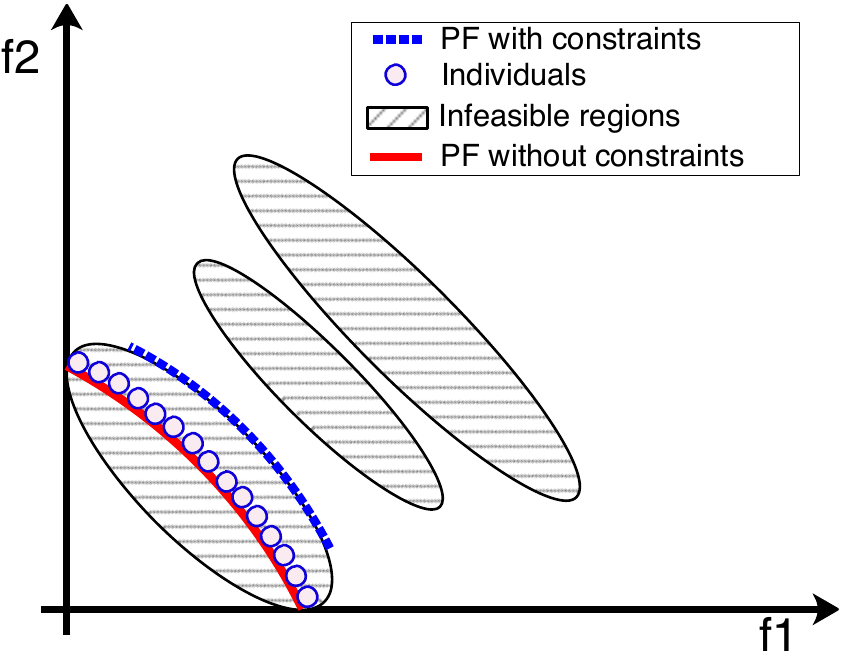}\\
\centering{\scriptsize{(c)}}
\end{minipage}
\end{tabular}

\vspace{0.5cm}
\begin{tabular}{cc}
\begin{minipage}[t]{0.28\linewidth}
\includegraphics[width = 5cm]{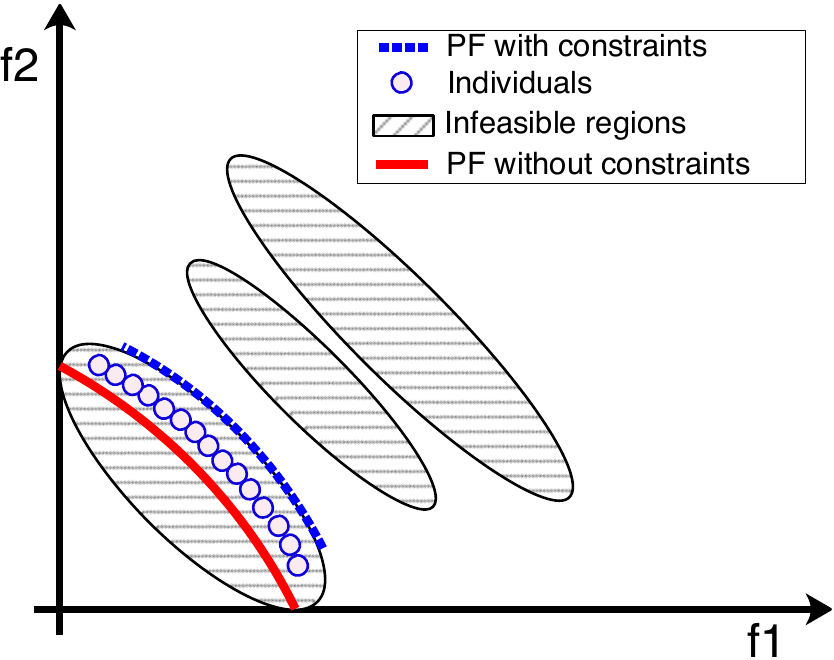}\\
\centering{\scriptsize{(d)}}
\end{minipage}
\hspace{0.5cm}
\begin{minipage}[t]{0.28\linewidth}
\includegraphics[width = 5cm]{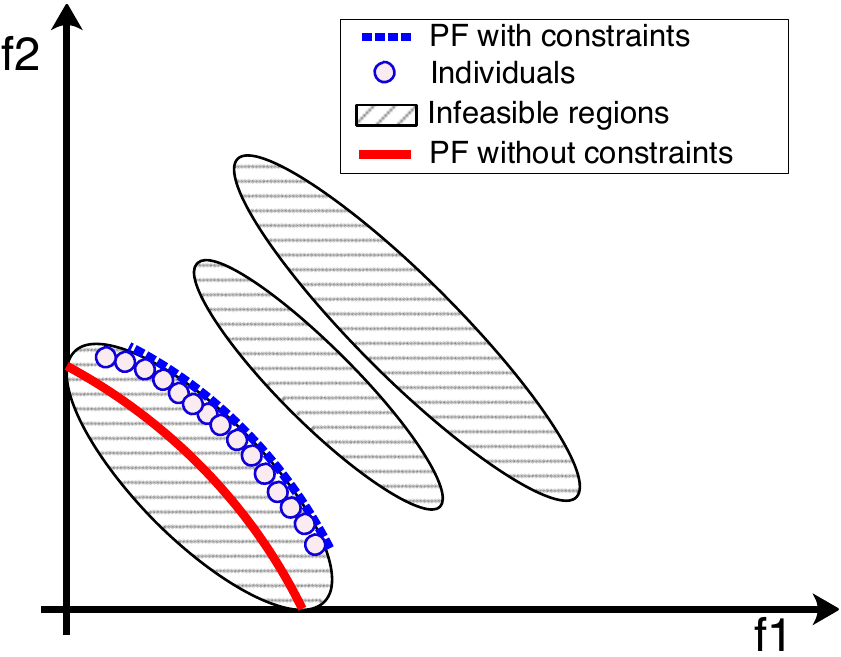}\\
\centering{\scriptsize{(e)}}
\end{minipage}
\hspace{0.5cm}
\begin{minipage}[t]{0.28\linewidth}
\includegraphics[width = 5cm]{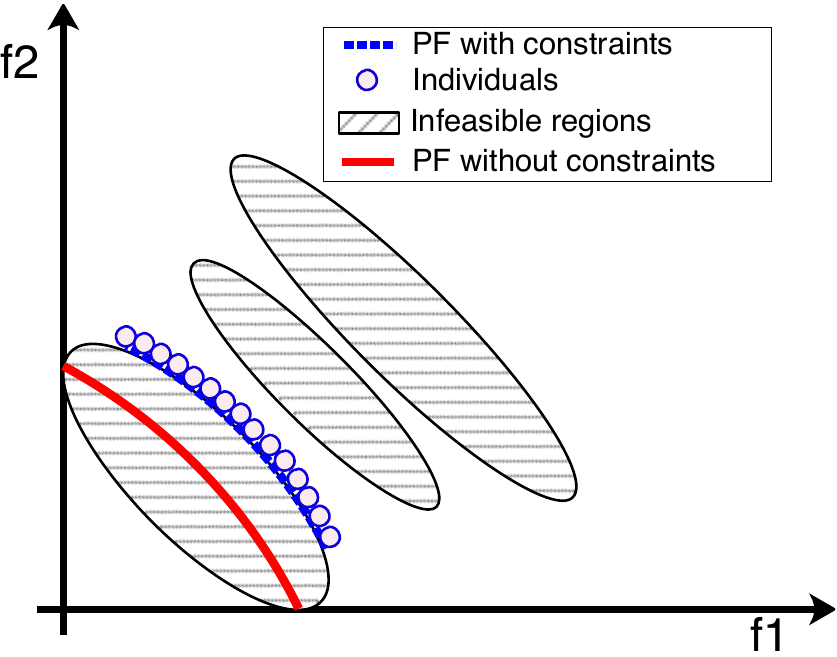}\\
\centering{\scriptsize{(f)}}
\end{minipage}
\end{tabular}
\caption{\label{fig:case-b} The unconstrained PF is covered by infeasible regions and becomes no more feasible. The real PF lies on some boundaries of constraints. (a)-(c) show the push search process, and the working population can get across the infeasible regions without any barriers. (e)-(f) show the pull search process, the infeasible solutions in the working population are pulled to the feasible and non-dominated regions gradually.}
\end{figure*}

\begin{figure*}
\begin{tabular}{cc}
\begin{minipage}[t]{0.28\linewidth}
\includegraphics[width = 5cm]{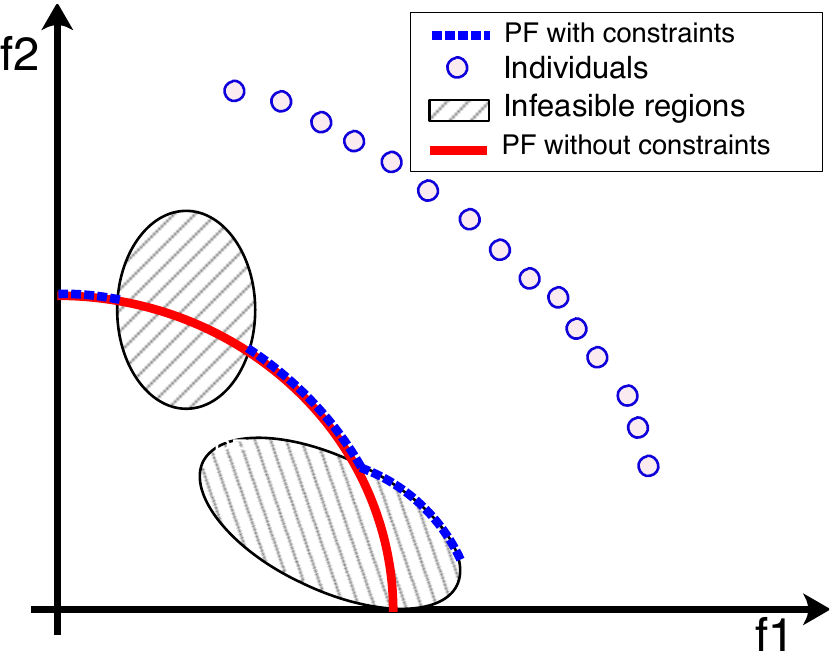}\\
\centering{\scriptsize{(a)}}
\end{minipage}
\hspace{0.5cm}
\begin{minipage}[t]{0.28\linewidth}
\includegraphics[width = 5cm]{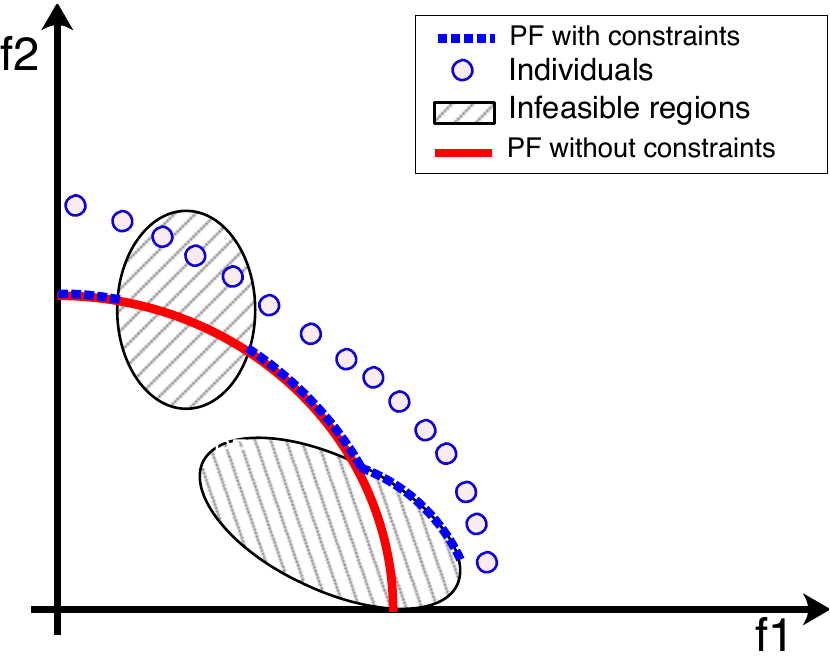}\\
\centering{\scriptsize{(b)}}
\end{minipage}
\hspace{0.5cm}
\begin{minipage}[t]{0.28\linewidth}
\includegraphics[width = 5cm]{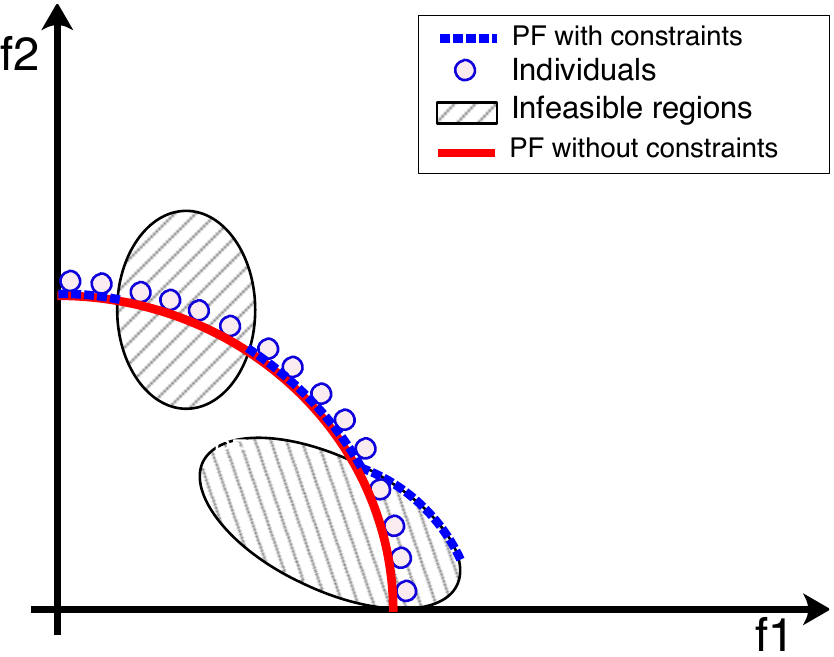}\\
\centering{\scriptsize{(c)}}
\end{minipage}
\end{tabular}

\vspace{0.5cm}
\begin{tabular}{cc}
\begin{minipage}[t]{0.28\linewidth}
\includegraphics[width = 5cm]{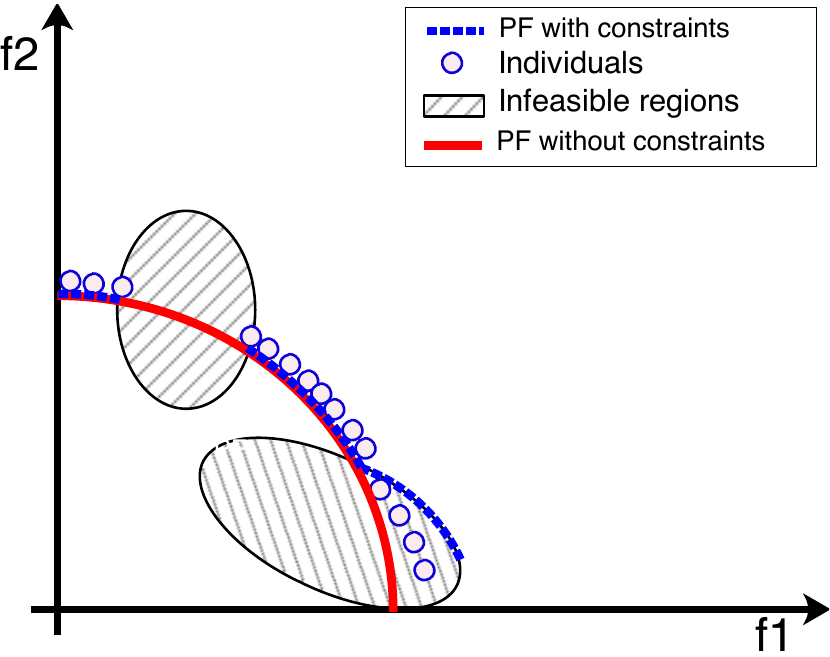}\\
\centering{\scriptsize{(d)}}
\end{minipage}
\hspace{0.5cm}
\begin{minipage}[t]{0.28\linewidth}
\includegraphics[width = 5cm]{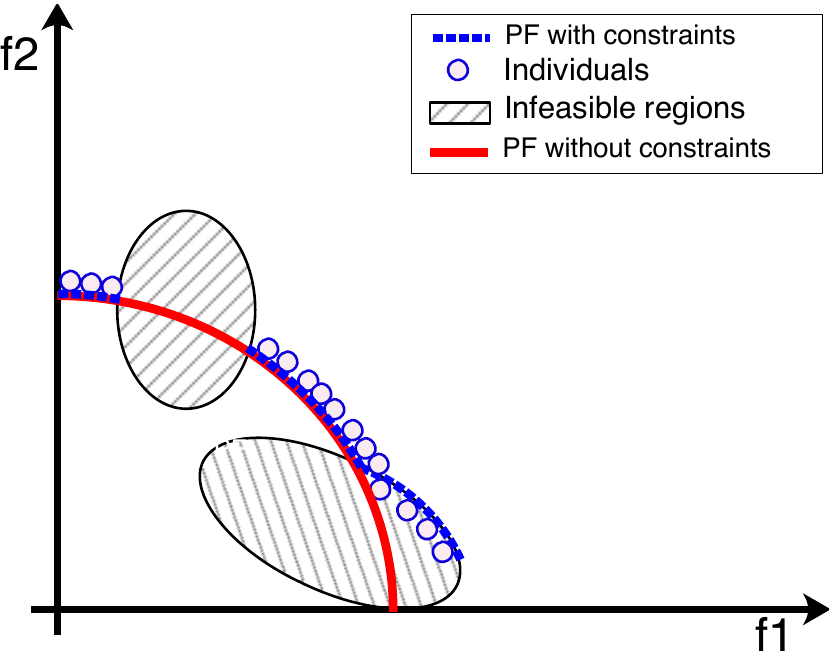}\\
\centering{\scriptsize{(e)}}
\end{minipage}
\hspace{0.5cm}
\begin{minipage}[t]{0.28\linewidth}
\includegraphics[width = 5cm]{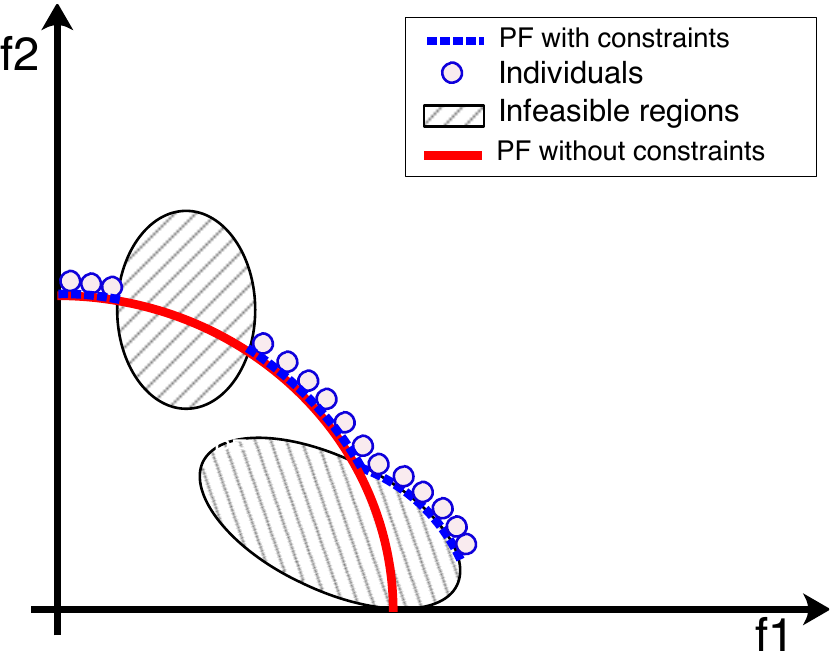}\\
\centering{\scriptsize{(f)}}
\end{minipage}
\end{tabular}
\caption{\label{fig:case-c} Infeasible regions make the original unconstrained PF partially feasible. (a)-(c) show the push search process, and some parts of the real PF have been found in this process. In the pull stage, infeasible solutions have been pulled to the feasible and non-dominated regions gradually as shown in (d)-(f).}
\end{figure*}

\begin{figure*}
\begin{tabular}{cc}
\begin{minipage}[t]{0.28\linewidth}
\includegraphics[width = 5cm]{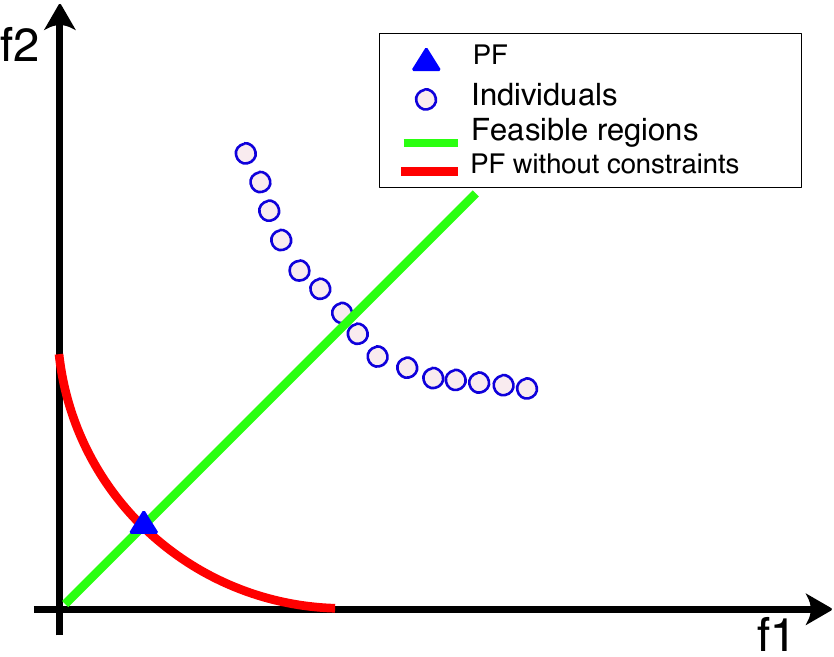}\\
\centering{\scriptsize{(a)}}
\end{minipage}
\hspace{0.5cm}
\begin{minipage}[t]{0.28\linewidth}
\includegraphics[width = 5cm]{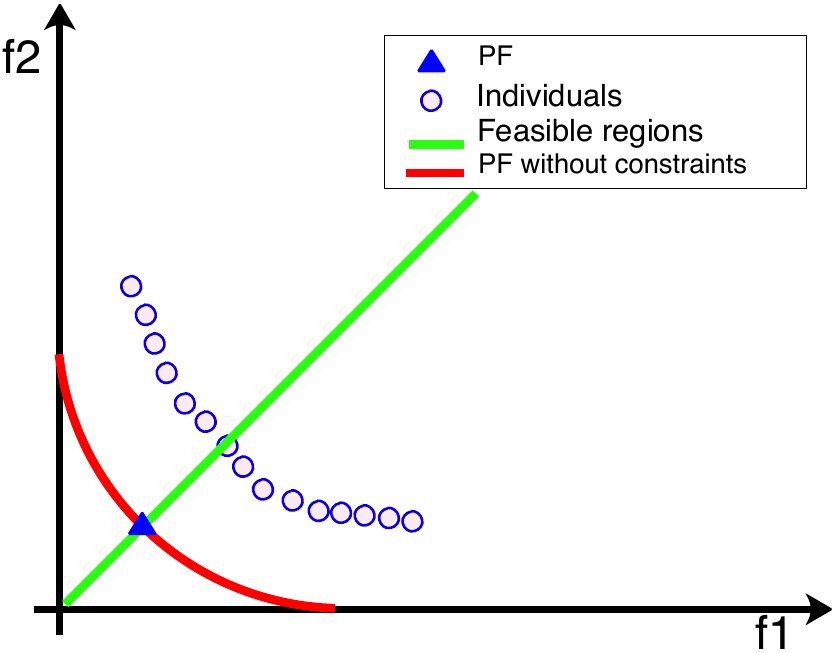}\\
\centering{\scriptsize{(b)}}
\end{minipage}
\hspace{0.5cm}
\begin{minipage}[t]{0.28\linewidth}
\includegraphics[width = 5cm]{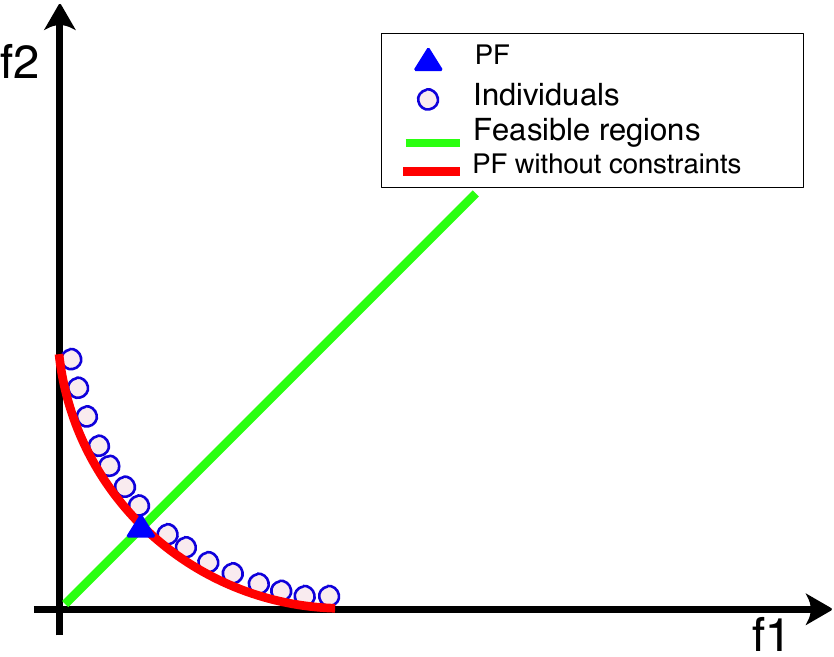}\\
\centering{\scriptsize{(c)}}
\end{minipage}
\end{tabular}

\vspace{0.5cm}
\begin{tabular}{cc}
\begin{minipage}[t]{0.28\linewidth}
\includegraphics[width = 5cm]{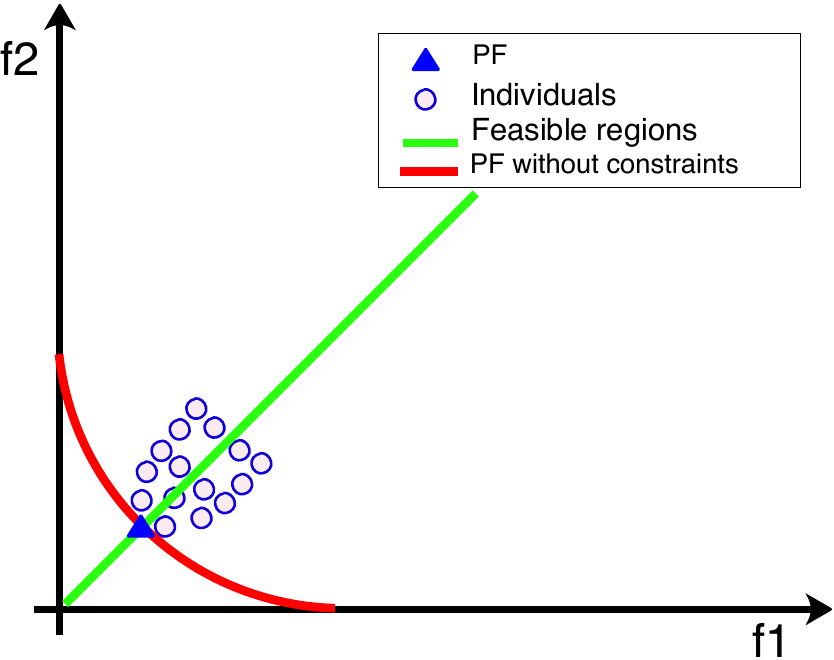}\\
\centering{\scriptsize{(d)}}
\end{minipage}
\hspace{0.5cm}
\begin{minipage}[t]{0.28\linewidth}
\includegraphics[width = 5cm]{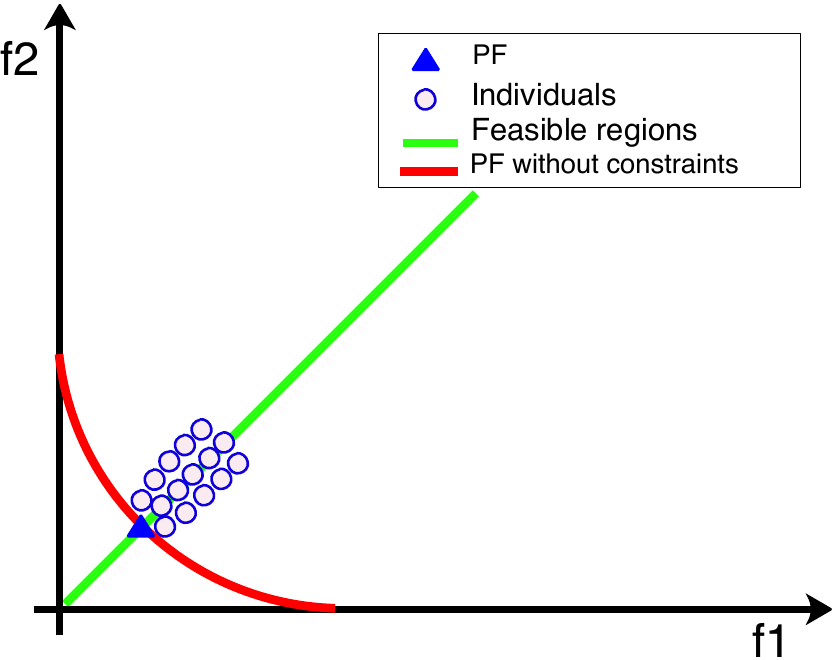}\\
\centering{\scriptsize{(e)}}
\end{minipage}
\hspace{0.5cm}
\begin{minipage}[t]{0.28\linewidth}
\includegraphics[width = 5cm]{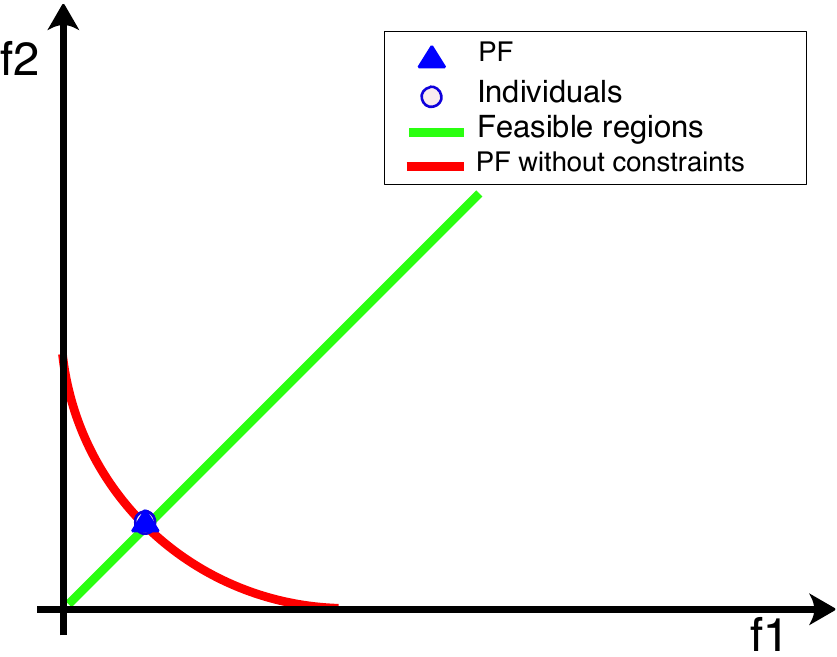}\\
\centering{\scriptsize{(f)}}
\end{minipage}
\end{tabular}
\caption{\label{fig:case-d} Infeasible regions reduce the dimensionality of the PF. (a)-(c) show the push search process of PPS, and a set of non-dominated solutions are achieved without considering any constraints. (d)-(f) show the pull search process of PPS, many infeasible solutions are pulled to the feasible and non-dominated regions, which reside in a reduced dimensionality of the space.}
\end{figure*}

Constraints define the infeasible regions in the decision space, which may also affect the PFs of CMOPs in the objective space. The influence of the infeasible regions on the PFs can be generally classified into four different situations. For each situation, the search behavior of PPS is illustrated by Fig. \ref{fig:case-a}-\ref{fig:case-d} respectively, which can be summarized as follows. 

\begin{enumerate}
\item Infeasible regions block the way towards the PF, as illustrated by Fig. \ref{fig:case-a}(a). In this circumstance, the unconstrained PF is the same as the constrained PF, and PPS has significant advantages compared with other CMOEAs. Since the constraints are ignored in the push stage of PPS, the infeasible regions have no effect on the searching of PPS. Fig. \ref{fig:case-a}(a)-(e) show the push process in different stages, where the working population crosses the infeasible regions in this case without any extra efforts. As the constrained PF is the same as the unconstrained PF, the true PF has already been approximated by the working population in the push process, the pull search has no affect on the working population, as shown in Fig. \ref{fig:case-a}(f).

\item The unconstrained PF is covered by infeasible regions and becomes no more feasible. Every constrained Pareto optimal point lies on some constraint boundaries, as illustrated by Fig. \ref{fig:case-b}(a). In this circumstance, PPS first approaches the unconstrained PF by using the push strategy as illustrated by Fig. \ref{fig:case-b}(a)-(c). After the working population approaches the unconstrained PF, the pull strategy is applied to pull the working population towards the real PF, as illustrated by Fig. \ref{fig:case-b}(d)-(f).  

\item Infeasible regions make the original unconstrained PF partially feasible, as illustrated by Fig. \ref{fig:case-c}(a). In this situation, some parts of the true PF have already been achieved during the push search as illustrated by Fig. \ref{fig:case-c}(c). In the pull stage, infeasible solutions are pulled to the feasible and non-dominated regions as illustrated by Fig. \ref{fig:case-c}(d)-(f). Finally, the whole real PF has been found by PPS. 

\item Infeasible regions may reduce the dimensionality of the PF of a CMOP, as illustrated by Fig. \ref{fig:case-d}(a). In the push stage, a set of non-dominated solutions without considering constraints are achieved as illustrated by Fig. \ref{fig:case-d}(a)-(c) . Then, infeasible solutions in the working population are pulled toward the feasible and non-dominated regions in the pull search process, as shown in Fig. \ref{fig:case-d}(d)-(f). 

\end{enumerate}

From the above analysis, it can be observed that PPS can deal with CMOPs with all situations of PFs affected by constraints.

The main steps of PPS includes the push and pull search processes. However, when to apply the push or pull search is very critical. A strategy to switch the search behavior is suggested as follows.

\begin{eqnarray}
\label{equ:r_rate}
r_k \equiv \max \{ rz_k, rn_k \} \le \epsilon
\end{eqnarray}
where $r_k$ represents the max rate of change between the ideal and nadir points during the last $l$-th generation. $\epsilon$ is a user-defined parameter, here we set $\epsilon = 1e-3$. The rates of change of the ideal and nadir points during the last $l$-th generation are defined in Eq. \ref{equ:z_rate} and Eq. \ref{equ:n_rate} respectively.  

\begin{eqnarray}
\label{equ:z_rate}
rz_k = \max_{i = 1,\ldots,m}\{  \frac{|z^k_i - z^{k-l}_i|}{ \max \{|z^{k-l}_i|,1e-6\}} \}
\end{eqnarray}

\begin{eqnarray}
\label{equ:n_rate}
rn_k = \max_{i = 1,\ldots,m}\{  \frac{|n^k_i - n^{k-l}_i|}{ \max \{|n^{k-l}_i|,1e-6\}} \}
\end{eqnarray}
where $z^k = (z^k_1,\ldots,z^k_m), n^k = (n^k_1,\ldots,n^k_m)$ are the ideal and nadir points in the $k$-th generation, and $z^k_i = \min_{j = 1,\ldots,N} f_i(\mathbf{x}^j), n^k_i = \max_{j = 1,\ldots,N} f_i(\mathbf{x}^j)$. $z^{k-l} = (z^{k-l}_1,\ldots,z^{k-l}_m), n^{k - l} = (n^{k-l}_1,\ldots,n^{k-l}_m)$ are the ideal and nadir points in the $(k-l)$-th generation.

At the beginning of the search, $r_k$ is initialized to $1.0$. At each generation, $r_k$ is updated according to Eq. \ref{equ:r_rate}. If $r_k$ is less or equal then the predefined threshold $\epsilon$, the search behavior is switched to the pull search. 

To summarize, PPS divides its search process into two different stages, including the push search and pull search. At the first stage, the push search without considering constraints is adopted to approximate the unconstrained PF. Once Eq. \ref{equ:r_rate} is satisfied, the pull search is used to pull the infeasible solutions to the feasible and non-dominated regions. In this situation, constraints are considered in the pull search process. PPS terminates when a predefined halting condition is met. In the following section, we will describe the instantiation of the push and pull strategy in a MOEA/D framework.

\section{An Instantiation of PPS in MOEA/D}
\label{sec:proposed-method}
In this section, the details of an instantiation of the push search method, as well as the method adopted in the pull search, and the proposed PPS in the framework of MOEA/D are described.

\subsection{The push search}
In the push search stage, an unconstrained MOEA/D is used to search for non-dominated solutions without considering any constraints. When solving a CMOP by using MOEA/D, we can decompose the CMOP into a set of single constrained optimization subproblems, and optimize them simultaneously in a collaborative way. Each subproblem is associated with a decomposition function by using a weight vector $\lambda ^ i$. In the decomposition-based selection approach, an individual is selected into next generation based on the value of the decomposition function. 

There are three popular decomposition approaches, including weighted sum \cite{miettinen1999nonlinear}, Tchebycheff \cite{miettinen1999nonlinear} and boundary intersection approaches \cite{Zhang:2007va}. In this paper, we adopt the Tchebycheff decomposition method, with the detailed definition given as follows.

\begin{eqnarray}
\label{equ:te_decom}
g^{te}(\mathbf{x},\lambda^i,z^*) = \max_{j = 1,\ldots,m} \frac{1}{\lambda^i_j}(|f_j(\mathbf{x}) - z^*_j|) 
\end{eqnarray}
where $\lambda^i$ is a weight vector, and $\sum_{j = 1,\ldots,m}{\lambda^i_j} = 1, \lambda_j^i \ge 0$. $z^*$ is the ideal points as described in section \ref{sec:pps}.  

In the push search stage, a newly generated solution $\mathbf{x}$ is kept to the next generation based on the value of $g^{te}$ as described in Algorithm \ref{alg:push}.

\begin{algorithm}
\Fn{result = PushSubproblems($\mathbf{x}^j$,$\mathbf{y}^j$)}{
    $result = false$\\
    \If{$g^{te}(\mathbf{y}^i|\lambda^j,z^{*}) \leq g^{te}(\mathbf{x}^j|\lambda^j,z^{*})$}{
            $\mathbf{x}^j$ = $\mathbf{y}^j$\\
            $result = ture$
        }
   \Return $result$
}
\caption{Push Subproblem}
\label{alg:push}
\end{algorithm}

\subsection{The pull search}
In this process, some infeasible solutions are pulled to the feasible and non-dominated regions. To achieve this, a constraint-handling mechanism is adopted to punish the infeasible solutions in the pull search stage. An improved epsilon constraint-handling to deal with constraints is proposed with the detailed formulation given as follows.

\begin{eqnarray}
&\varepsilon(k) = \begin{cases}
(1 - \tau) \varepsilon(k-1), \text{if }rf_k < \alpha\\
\varepsilon(0)(1 - \frac{k}{T_c})^{cp}, \text{if }rf_k \ge \alpha\\
\end{cases}
\label{equ:iepsilon_setting}
\end{eqnarray}
where $rf_k$ is the ratio of feasible solutions in the $k$-th generation. $\tau$ is to control the speed of reducing the relaxation of constraints in the case of $rf_k < \alpha$, and $\tau \in [0,1]$. $\alpha$ is to control the searching preference between the feasible and infeasible regions, and $\alpha \in[0,1]$. $cp$ is to control the speed of reducing relaxation of constraints in the case of $rf_k \ge \alpha$. $\varepsilon(k)$ is updated until the generation counter $k$ reaches the control generation $T_c$. $\varepsilon(0)$ is set to the maximum overall constraint violation of the working population at the end of the push search. 

The above proposed epsilon setting method in Eq. \ref{equ:iepsilon_setting} is different from our previous work \cite{wenji2016ssci} and \cite{fan2017improved}. In this paper, $\varepsilon(0)$ is set to the maximum overall constraint violation of the working population at the end of the push search, while $\varepsilon(0)$ is set to the overall constraint violation of the top $\theta$-th individual in the initial population in \cite{wenji2016ssci} and \cite{fan2017improved}. In the case of $rf_0 == 0$, $\varepsilon(0) = 0$ according to the epsilon setting in \cite{wenji2016ssci} and \cite{fan2017improved}, which can not help to explore the infeasible regions and get across large infeasible regions. To avoid this situation, $\varepsilon(k)$ is increased if $rf_k$ is greater than a predefined threshold in \cite{wenji2016ssci} and \cite{fan2017improved}. However, increasing $\varepsilon(k)$ properly is still an issue that needs to be further investigated. In this paper, $\varepsilon(k)$ does not need to be increased to get across large infeasible regions, because the working population has already gotten across large infeasible regions in the push stage.


In the pull stage, a newly generated solution $\mathbf{x}$ is selected into the next generation based on the value of $g^{te}$ and the overall constraint violation $\phi(\mathbf{x})$ as illustrated by Algorithm \ref{alg:pull}.

\begin{algorithm}
\Fn{result = PullSubproblems($\mathbf{x}^j$,$\mathbf{y}^j$,$\varepsilon(k)$)}{
    $result = false$

    \uIf{$\phi(\mathbf{y}^j) \le \varepsilon(k)$ and $\phi(\mathbf{x}^j) \le \varepsilon(k)$}{
        \If{$g^{te}(\mathbf{y}^i|\lambda^j,z^{*}) \leq g^{te}(\mathbf{x}^j|\lambda^j,z^{*})$}{
            $\mathbf{x}^j$ = $\mathbf{y}^j$\\
            $result = ture$
        }
    }
   \uElseIf{$\phi(\mathbf{y}^j) == \phi(\mathbf{x}^j)$}{
        \If{$g^{te}(\mathbf{y}^j|\lambda^j,z^{*}) \leq g^{te}(\mathbf{x}^j|\lambda^j,z^{*})$}{
            $\mathbf{x}^j$ = $\mathbf{y}^j$\\
            $result = ture$
        }
   }
   \ElseIf{$\phi(\mathbf{y}^j) < \phi(\mathbf{x}^j)$}{
        $\mathbf{x}^j$ = $\mathbf{y}^j$\\
        $result = ture$
    }

    \Return $result$
}
\caption{Pull Subproblem}
\label{alg:pull}
\end{algorithm}

\subsection{PPS Embedded in MOEA/D}
Algorithm \ref{alg:pps} outlines the psuecode of PPS-MOEA/D, a PPS embedded in MOEA/D. A CMOP is decomposed into $N$ single objective subproblems, and these subproblems are initialized at line 1. The max rate of change of ideal and nadir points is initialized to $1.0$, and the flag of search stage $PushStage = true$. Then, the algorithm repeats lines 3-27 until $T_{max}$ generations have been reached. The value of $\varepsilon(k)$ and the search strategy are set at lines 4-14 based on Eq. \ref{equ:r_rate} and Eq. \ref{equ:iepsilon_setting}. $\varepsilon(k)$ and $\varepsilon(0)$ are initialized at line 7, and the function MaxViolation() calculates the maximum overall constraint violation in the current working population. Lines 15-19 show the process of generating a new solution. The updating of the ideal points are illustrated by lines 20-22. Lines 23-33 show the updating of subproblems, and different search strategies are adopted according to the state of the searching. In the case of $PushStage == true$, the push search is adopted, otherwise, the pull search is used. The generation count $k$ is updated at line 35. The set of feasible and non-dominated solutions $NS$ is updated according to the non-dominated ranking in NSGA-II \cite{996017} at line 36.

\emph{Remark:} The proposed PPS is a general framework for solving CMOPs. It can be instantiated in many different MOEAs, even though only PPS-MOEA/D is realized in this paper. At each search stage, many information can be gathered to extract useful knowledge which can be used to conduct the searching in both search stages. In fact, knowledge discovery is a critical step in the PPS framework. In this paper, we only adopts some statistic information. For example, the maximum overall constraint violation at the end of the push search stage is adopted to set the value of $\varepsilon(0)$. The ratio of feasible solutions at the pull search stage is used to control the value of $\varepsilon(k)$. In fact, many data mining methods and machine learning approaches can be integrated in the PPS framework for solving CMOPs more efficiently.

\begin{algorithm}
    \KwIn{$N$: the number of subproblems. $T_{max}$: the maximum generation. $N$ weight vectors: $\mathbf{\lambda}^1,\ldots,\mathbf{\lambda}^N$. $T$: the size of the neighborhood. $\delta$: the selecting probability from neighbors. $n_r$: the maximal number of solutions replaced by a child.\\
    }
    \KwOut{$NS:$ a set of feasible non-dominated solutions}

    Decompose a CMOP into $N$ subproblems associated with $\mathbf{\lambda}^1,\ldots,\mathbf{\lambda}^N$. Generate an initial population $P=\{\mathbf{x}^1, \ldots, \mathbf{x}^N \}$.
    For each $i = 1, \dots, N$, set $B(i) = \{i_1,\dots,i_T\}$, where $\mathbf{\lambda}^{i_1},\dots,\mathbf{\lambda}^{i_T}$ are the $T$ closest weight vectors to $\mathbf{\lambda}^i$.\\
    Initialize the rate of change of ideal and nadir points ($maxChange = 1.0$) over the previous $L$ generations. Set $PushStage = true$, $k = 1$.\\

    \While{$k \le T_{max}$}{
    
    \lIf{$k >= L$}{
         $maxChange$ = CalcMaxChange($k$) according to Eq. \ref{equ:r_rate}
    }

    \eIf{$k < T_c$}{
    \If{$maxChange \le \alpha$ And PushStage == $true$}{
         PushStage = $false$; $\varepsilon(k)$ = $\varepsilon(0)$ = MaxViolation();
         }

         \If{$PushStage == false$}{
             Update $\varepsilon(k)$ according to Eq. \ref{equ:iepsilon_setting};
         }
    }{
       $\varepsilon(k)$ = 0;
    }

    Generate a random permutation $rp$ from $\{1,\ldots,N\}$.\\
    \For{$i \leftarrow 1$ \KwTo $N$}{
    Generate a random number $r\in[0,1]$, $j = rp(i)$.\\
    \leIf{$r < \delta$}{
    $S = B(j)$
    }{
    $ S = \{1,\ldots,N\}$
    }
    Generate $\mathbf{y}^j$ through the DE operator, and perform polynomial mutation on $\mathbf{y}^j$.\\

    \For{$t \leftarrow 1$ \KwTo $m$}{
    \lIf{$z^*_t > f_t(\mathbf{y}^j)$}{
        $z^*_t  = f_t(\mathbf{y}^j)$
    }
    }
    Set $c = 0$.\\
    \While{$c \neq n_r$ or $S \neq \varnothing$}{
     select an index $j$ from $S$ randomly.\\
     \eIf{PushStage == $true$}{
      $result = PushSubproblems(\mathbf{x}^j, \mathbf{y}^j)$;// Algorithm \ref{alg:push};
    }{
      $result = PullSubproblems(\mathbf{x}^j, \mathbf{y}^j,\varepsilon(k))$; // Algorithm \ref{alg:pull};
    }
     \lIf{$result == true$}{
     $c = c+1$}
     $S = S \backslash\{j\}$;
    }

    }
    $k = k + 1$;\\
    $NS$ = NDSelect($NS  \bigcup P$); // According to NSGA-II \cite{996017};
    }
\caption{PPS-MOEA/D}
\label{alg:pps}
\end{algorithm}



\section{Experimental study}
\label{sec:exper}

\subsection{Experimental Settings}
To evaluate the performance of the proposed PPS method, the other five CMOEAs, including MOEA/D-IEpsilon, MOEA/D-Epsilon, MOEA/D-SR, MOEA/D-CDP and C-MOEA/D, are tested on LIR-CMOP1-14 \cite{fan2017improved}. The detailed parameters are listed as follows:

\begin{enumerate}
\item The mutation probability $Pm = 1/n$ ($n$ denotes the dimension of a decision vector). The distribution index in the polynomial mutation is set to 20.
\item DE parameters: $CR = 1.0$, $f = 0.5$.
\item Population size: $N = 300$. Neighborhood size: $T = 30$.
\item Halting condition: each algorithm runs for 30 times independently, and stops until 300,000 function evaluations are reached.
\item Probability of selecting individuals from its neighborhood: $\delta = 0.9$.
\item The max number of solutions updated by a child: $nr = 2$.
\item Parameters setting in PPS-MOEA/D: $T_c = 800$, $\alpha = 0.95$, $\tau = 0.1$, $cp = 2$, $l = 20$.
\item Parameters setting in MOEA/D-IEpsilon: $T_c = 800$, $\alpha = 0.95$, $\tau = 0.1$, and $\theta = 0.05 N$. 
\item Parameters setting in MOEA/D-Epsilon: $T_c = 800$, $cp = 2$, and $\theta = 0.05 N$.
\item Parameter setting in MOEA/D-SR: $S_r = 0.05$.
\end{enumerate}

\subsection{Performance Metric}

To measure the performance of PPS-MOEA/D, MOEA/D-IEpsilon, MOEA/D-Epsilon, MOEA/D-CDP, MOEA/D-SR and C-MOEA/D, two popular metrics--the inverted generation distance (IGD) \cite{bosman2003balance} and the hypervolume \cite {Zitzler1999Multiobjective} are adopted.
\begin{itemize}
\item \textbf{Inverted Generational Distance} (IGD):
\end{itemize}
The IGD metric reflects the performance regarding convergence and diversity simultaneously. The detailed definition is given as follows:
\begin{equation} \label{IGD metric}
\begin{cases}
IGD(P^*,A) = \frac{\sum \limits_{y^* \in P^*}d(y^*,A)}{| P^* |}\\
\\
d(y^*,A) = \min \limits_{y \in A} \{ \sqrt {\sum_{i = 1} ^m (y^{*}_{i} - y_i)^2} \}
\end{cases}
\end{equation}
where $P^*$ denotes a set of representative solutions in the real PF, $A$ is an approximate PF achieved by a CMOEA. $m$ denotes the number of objectives. For LIR-CMOPs with two objectives, 1000 points are sampled uniformly from the true PF to construct $P^*$. For LIR-CMOPs with three objectives, 10000 points are sampled uniformly from the PF to constitute $P^*$. It is worth noting that a smaller value of IGD may indicate better performance with regards to diversity and/or convergence.


\begin{itemize}
\item \textbf{Hypervolume} ($HV$):
\end{itemize}
$HV$ reflects the closeness of the set of non-dominated solutions achieved by a CMOEA to the real PF. The larger $HV$ means that the corresponding non-dominated set is closer to the true PF. 

\begin{equation}
HV(S)=VOL(\bigcup \limits_{x\in S} [f_1(x),z_1^r]\times ...[f_m(x),z_m^r] ) \\
\end{equation}
where $VOL(\cdot)$ is the Lebesgue measure, $m$ denotes the number of objectives, $\mathbf{z}^r=(z_1^r,...,z_m^r)^T$ is a user-defined reference point in the objective space. For each LIR-CMOP, the reference point is placed at 1.2 times the distance to the nadir point of the true PF. It is worth noting that a larger value of $HV$ may indicate better performance regarding diversity and/or convergence. 

\subsection{Discussion of Experiments}

\subsubsection{A Comparison between PPS-MOEA/D and MOEA/D-IEpsilon}

The difference between PPS-MOEA/D and MOEA/D-IEpsilon is that MOEA/D-IEpsilon does not apply the push search without considering any constraints, and they adopt different epsilon setting methods. To illustrate the effect of the PPS framework, the performance regarding IGD and HV metrics of PPS-MOEA/D and MOEA/D-IEpsilon on LIR-CMOP1-14 is compared.

The statistical results of the IGD values on LIR-CMOP1-14 achieved by PPS-MOEA/D and MOEA/D-IEpsilon in 30 independent runs are listed in Table \ref{tab:igd}. For LIR-CMOP1-2, LIR-CMOP5-8, LIR-CMOP11 and LIR-CMOP14, PPS-MOEA/D performs significantly better than MOEA/D-IEpsilon. For LIR-CMOP9, MOEA/D-IEpsilon is significantly better than PPS-MOEA/D. For the rest of test instances, MOEA/D-IEpsilon and PPS-MOEA/D have no significant difference. 

The statistical results of the HV values on LIR-CMOP1-14 achieved by PPS-MOEA/D and MOEA/D-IEpsilon in 30 independent runs are listed in Table \ref{tab:hv}. For LIR-CMOP1-3, LIR-CMOP5-8 and LIR-CMOP14, PPS-MOEA/D is significantly better than MOEA/D-IEpsilon. There is no significantly difference between PPS-MOEA/D and MOEA/D-IEpsilon on LIR-CMOP4 and LIR-CMOP10-13. For LIR-CMOP9, PPS is significantly worse than MOEA/D-IEpsilon.

\begin{figure*}
\begin{tabular}{cc}
\begin{minipage}[t]{0.28\linewidth}
\includegraphics[width = 5.5cm]{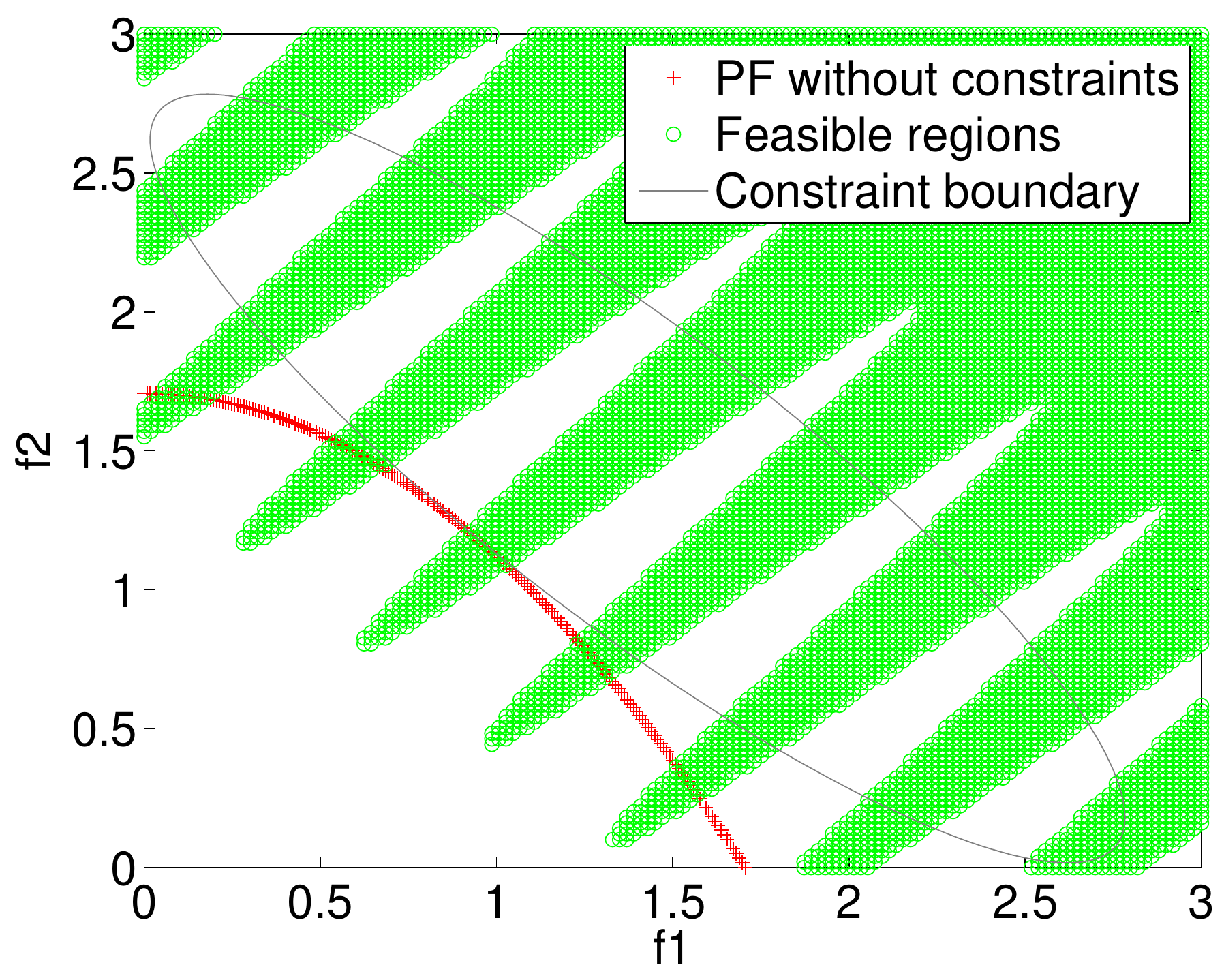}\\
\centering{\scriptsize{(a) The landscape of LIR-CMOP9}}
\end{minipage}
\hspace{0.5cm}
\begin{minipage}[t]{0.28\linewidth}
\includegraphics[width = 6.0cm]{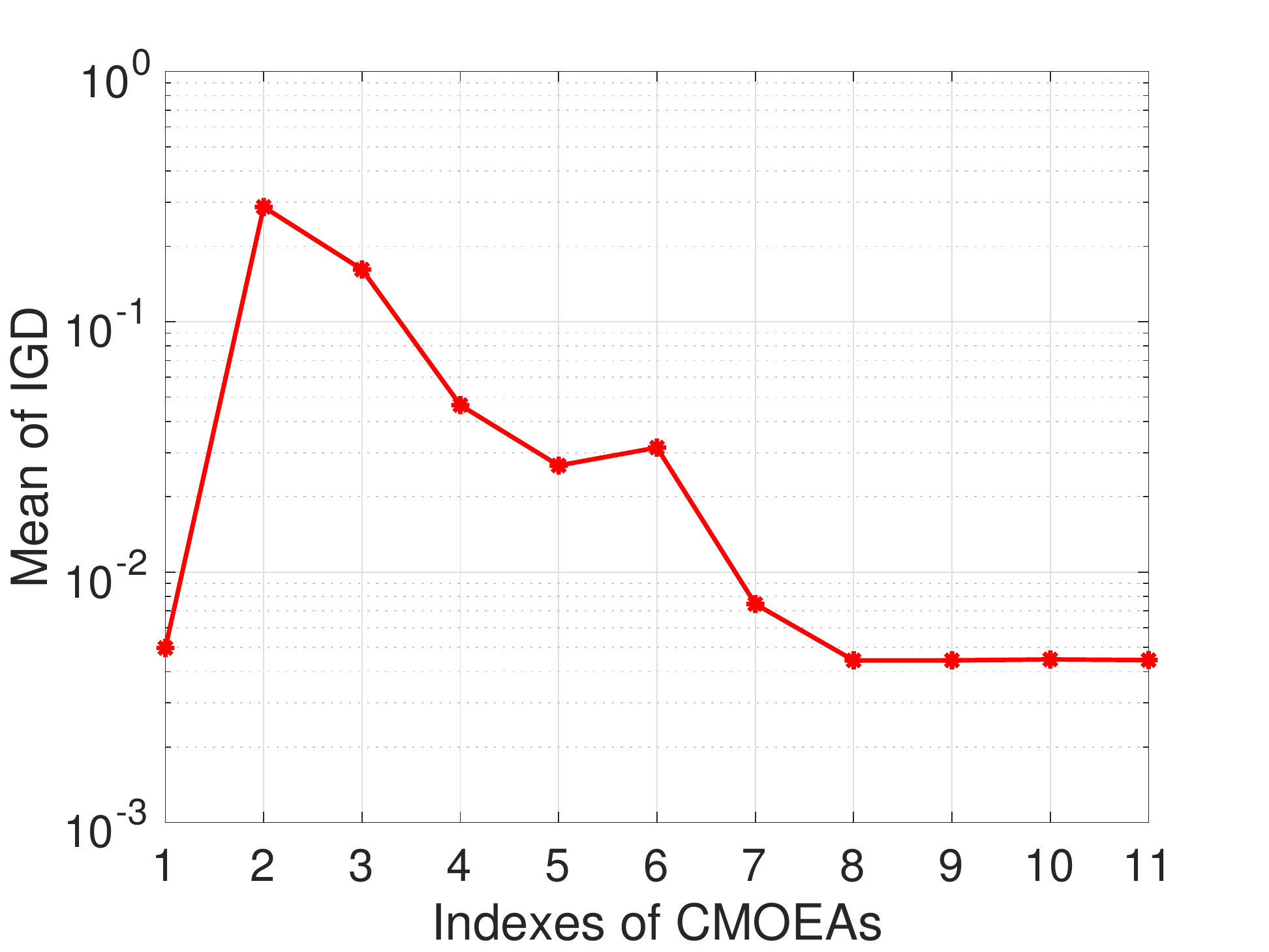}\\
\centering{\scriptsize{(b) IGD}}
\end{minipage}
\hspace{0.5cm}
\begin{minipage}[t]{0.28\linewidth}
\includegraphics[width = 6.0cm]{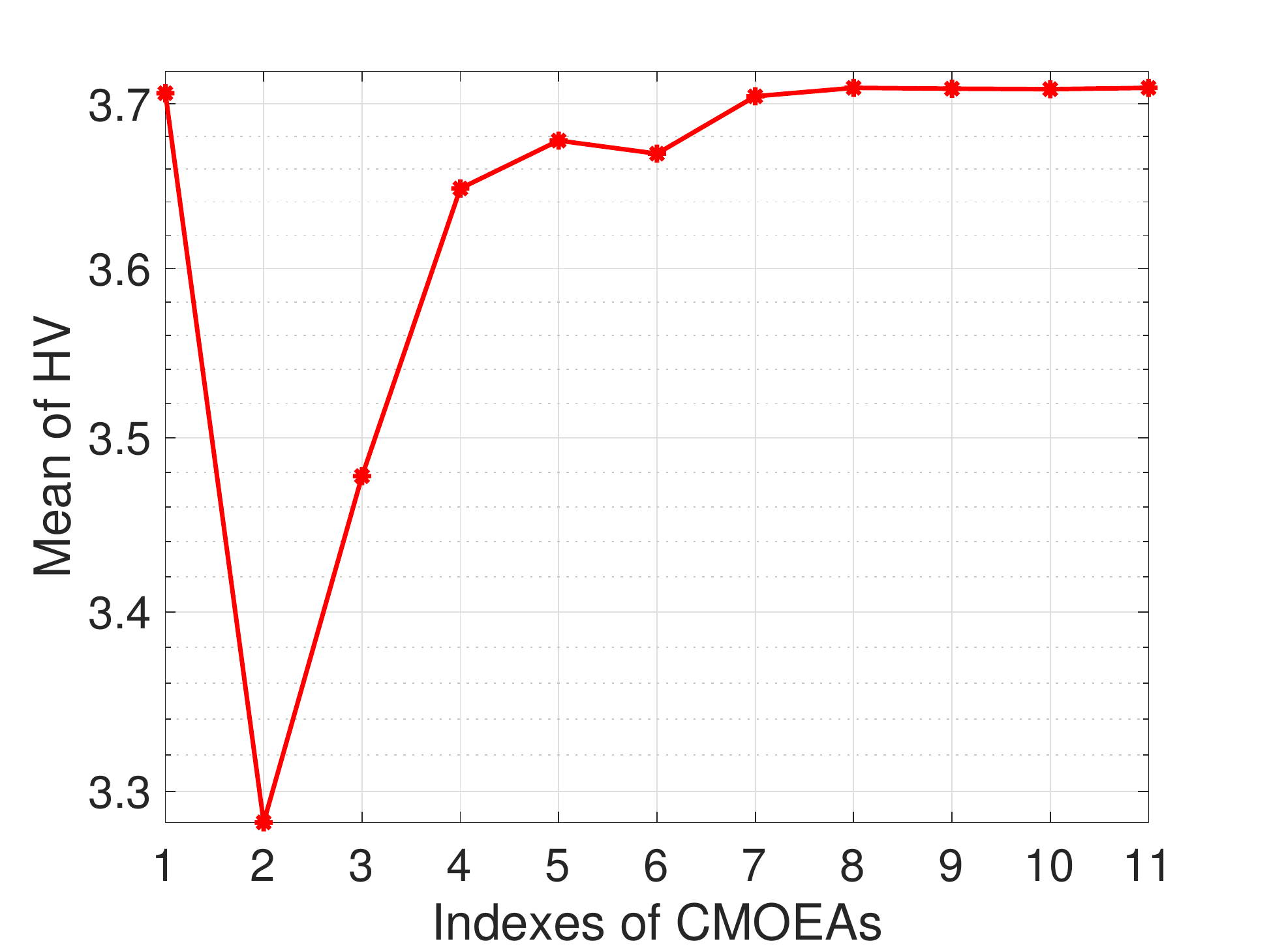}\\
\centering{\scriptsize{(c) HV}}
\end{minipage}
\end{tabular}

\caption{\label{fig:lircmop9} The feasible and infeasible regions of LIR-CMOP9 are plotted in (a). In (b)-(c), the mean values of IGD and HV of MOEA/D-IEpsilon and PPS-MOEA/D with different values of $l$ are plotted. In the horizontal axises of (b)-(c), the index '1' represents MOEA/D-IEpsilon, indexes '2-11' denote PPS-MOEA/D with $l = 10, 20, 30 ,40, 50,  60, 70, 80, 90$ and $100$.}
\end{figure*}

Since the real PF of LIR-CMOP9 locates on its unconstrained PF and has several disconnected parts, as illustrated by Fig. \ref{fig:lircmop9}(a), PPS-MOEA/D should performs well on this problems. However, the statistic results of both IGD and HV indicate that PPS-MOEA/D is significantly worse than MOEA/D-IEpsilon. One possible reason is that PPS-MOEA/D has not converged to the whole unconstrained PF in the push search stage.

To verify this hypothesis, we adjust the value of $l$ in Eq. \eqref{equ:z_rate} and Eq. \eqref{equ:n_rate}, which is a critical parameter to switch the search behavior. The mean values of IGD and HV values of MOEA/D-IEpsilon and PPS-MOEA/D with different values of $l$ are shown in Fig. \ref{fig:lircmop9}(b) and Fig. \ref{fig:lircmop9}(c), respectively. We can observe that, in cases of $l \ge 70$, PPS-MOEA/D is better than MOEA/D-IEpsilon.

To summarize, PPS-MOEA/D is superior to MOEA/D-IEpsilon on eight test problems, is not significantly different from MOEA/D-IEpsilon on five test problems, is inferior to MOEA/D-IEpsilon on only one test instance. It can be concluded that the PPS framework can help to enhance the performance of a CMOEA in terms of the IGD and HV metrics.

\subsubsection{A Comparison of PPS-MOEA/D and the other four decomposition-based CMOEAs}
The statistical results of the IGD values on LIR-CMOP1-14 achieved by PPS-MOEA/D and the other four decomposition-based CMOEAs in 30 independent runs are listed in Table \ref{tab:igd}. According to the Wilcoxon-Test in this table, it is clear that PPS-MOEA/D is significantly better than MOEA/D-Epsilon, MOEA/D-CDP, MOEA/D-SR, C-MOEA/D on all of the fourteen tested problems in terms of the IGD metric.

The statistical results of the HV values on LIR-CMOP1-14 achieved by PPS-MOEA/D and the other four decomposition-based CMOEAs in 30 independent runs are listed in Table \ref{tab:hv}. It can be observed that PPS-MOEA/D is significantly better than MOEA/D-Epsilon, MOEA/D-CDP, MOEA/D-SR and C-MOEA/D on all the test instances in terms of the HV metric. It is clear that the proposed PPS-MOEA/D performs better than the other four decomposition-based CMOEAs on all of the test instances.   

\subsubsection{The advantages of PPS-MOEA/D}
\begin{figure*}
\begin{tabular}{cc}
\begin{minipage}[t]{0.28\linewidth}
\includegraphics[width = 6.0cm]{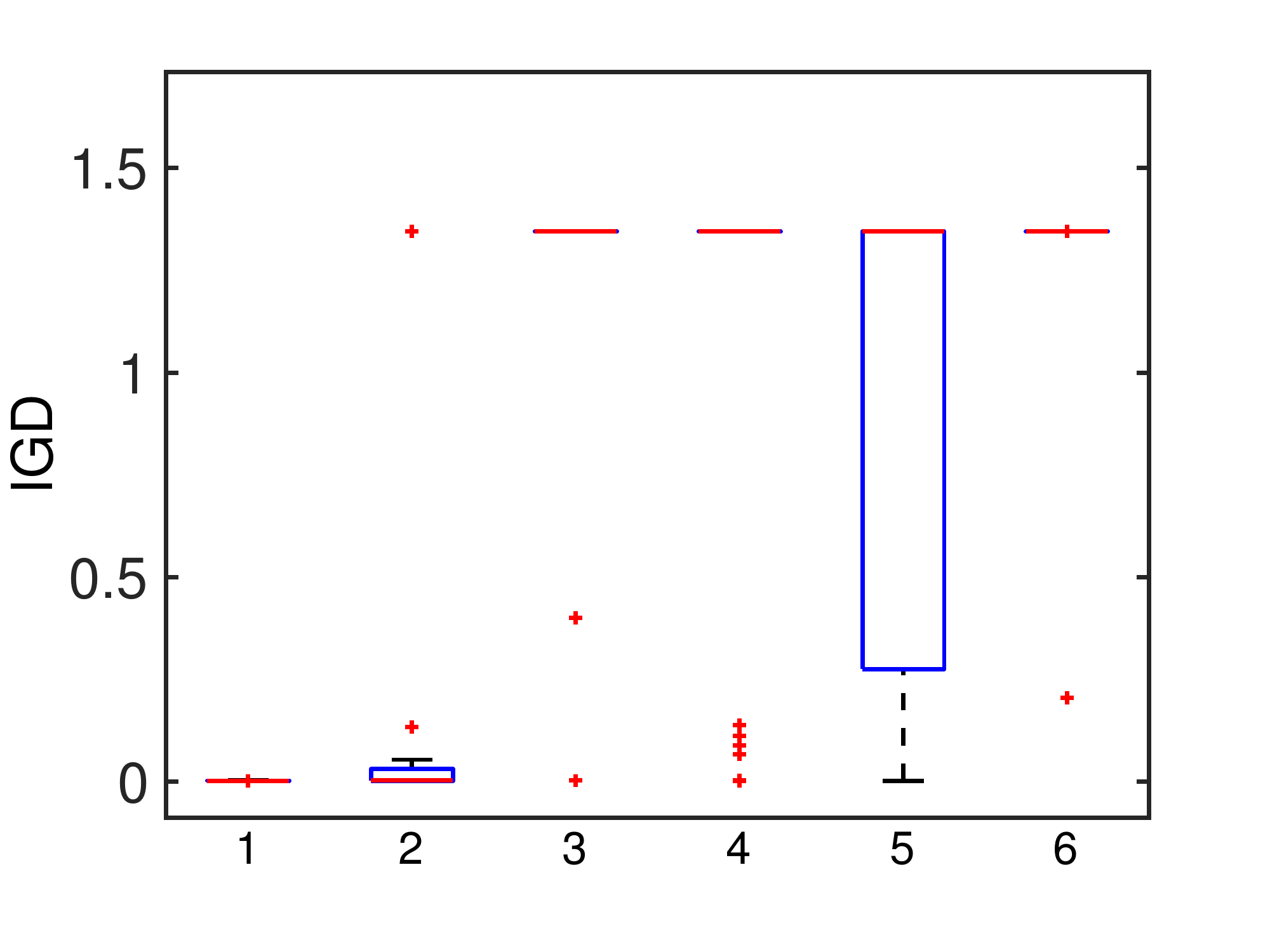}\\
\centering{\scriptsize{(a) LIR-CMOP6}}
\end{minipage}
\hspace{0.5cm}
\begin{minipage}[t]{0.28\linewidth}
\includegraphics[width = 6.0cm]{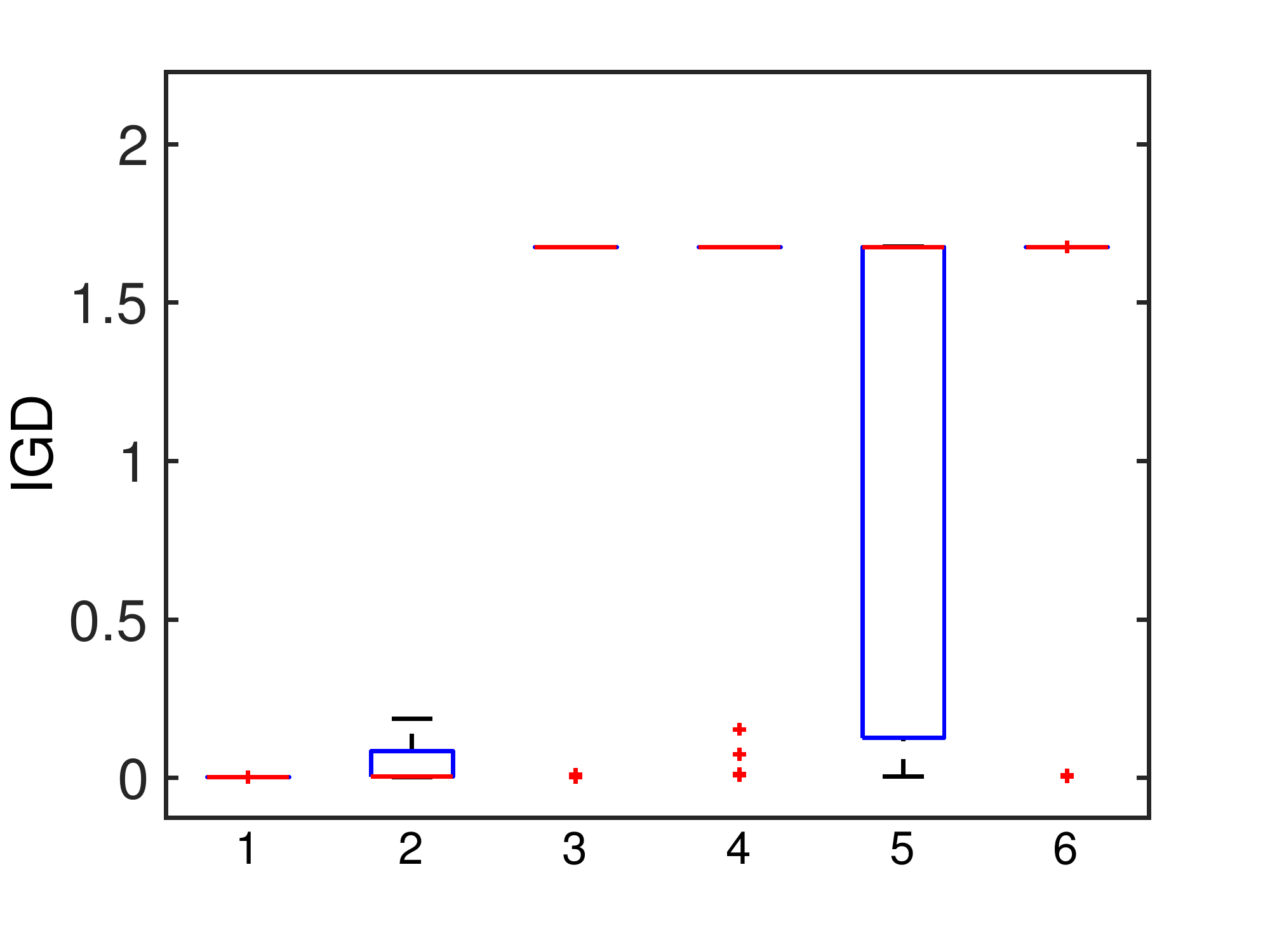}\\
\centering{\scriptsize{(b) LIR-CMOP7}}
\end{minipage}
\hspace{0.5cm}
\begin{minipage}[t]{0.28\linewidth}
\includegraphics[width = 6.0cm]{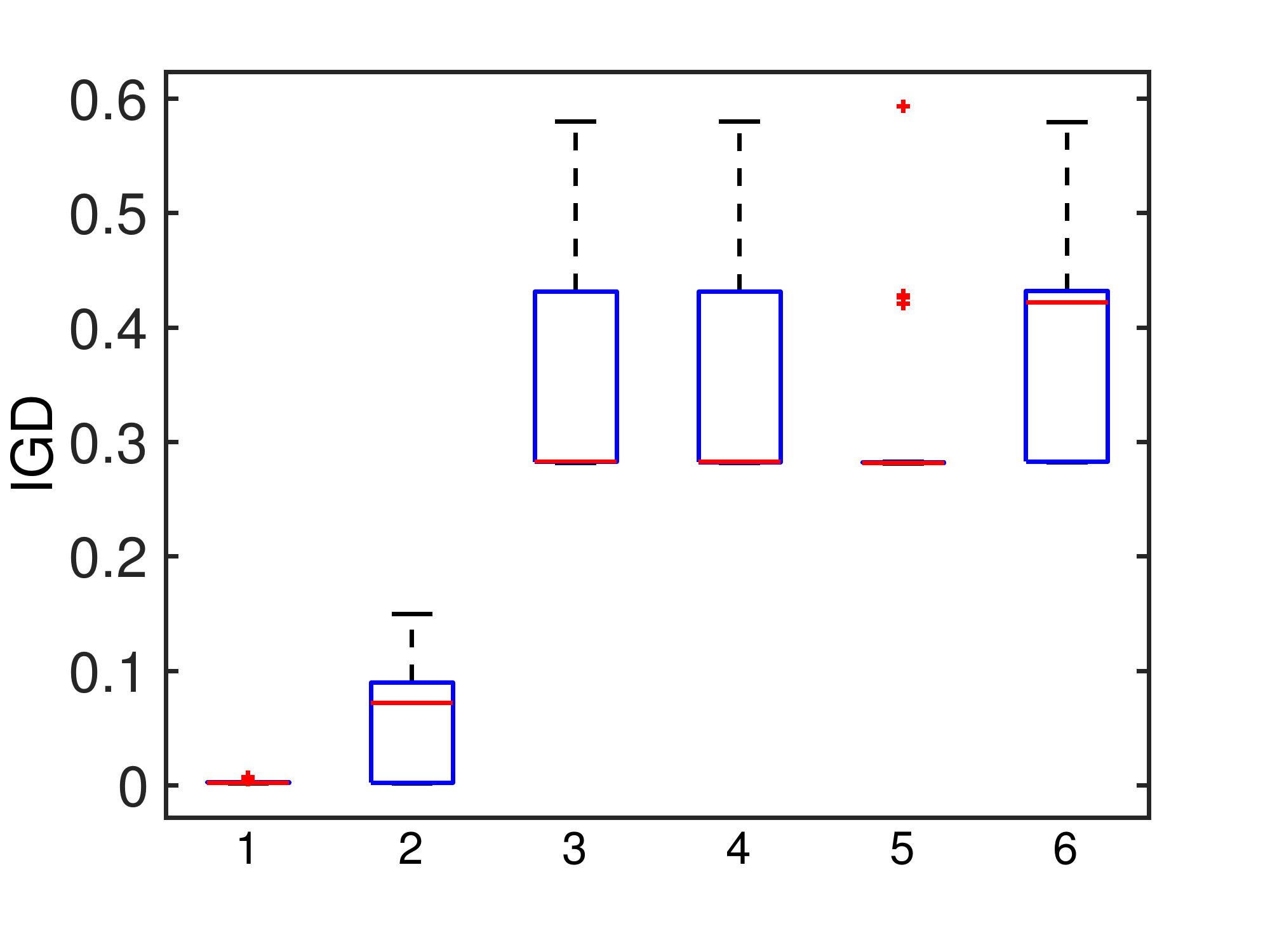}\\
\centering{\scriptsize{(c) LIR-CMOP11}}
\end{minipage}
\end{tabular}

\caption{\label{fig:igd-box} The box plots of IGD values for the six tested CMOEAs. The horizontal axis shows the index of each tested CMOEAs, where 1, 2, 3, 4, 5 and 6 denote PPS-MOEA/D, MOEA/D-IEpsilon, MOEA/D-Epsilon, MOEA/D-CDP, MOEA/D-SR and C-MOEA/D respectively.}
\end{figure*}

\begin{figure*}
\begin{tabular}{cc}
\begin{minipage}[t]{0.28\linewidth}
\includegraphics[width = 6.0cm]{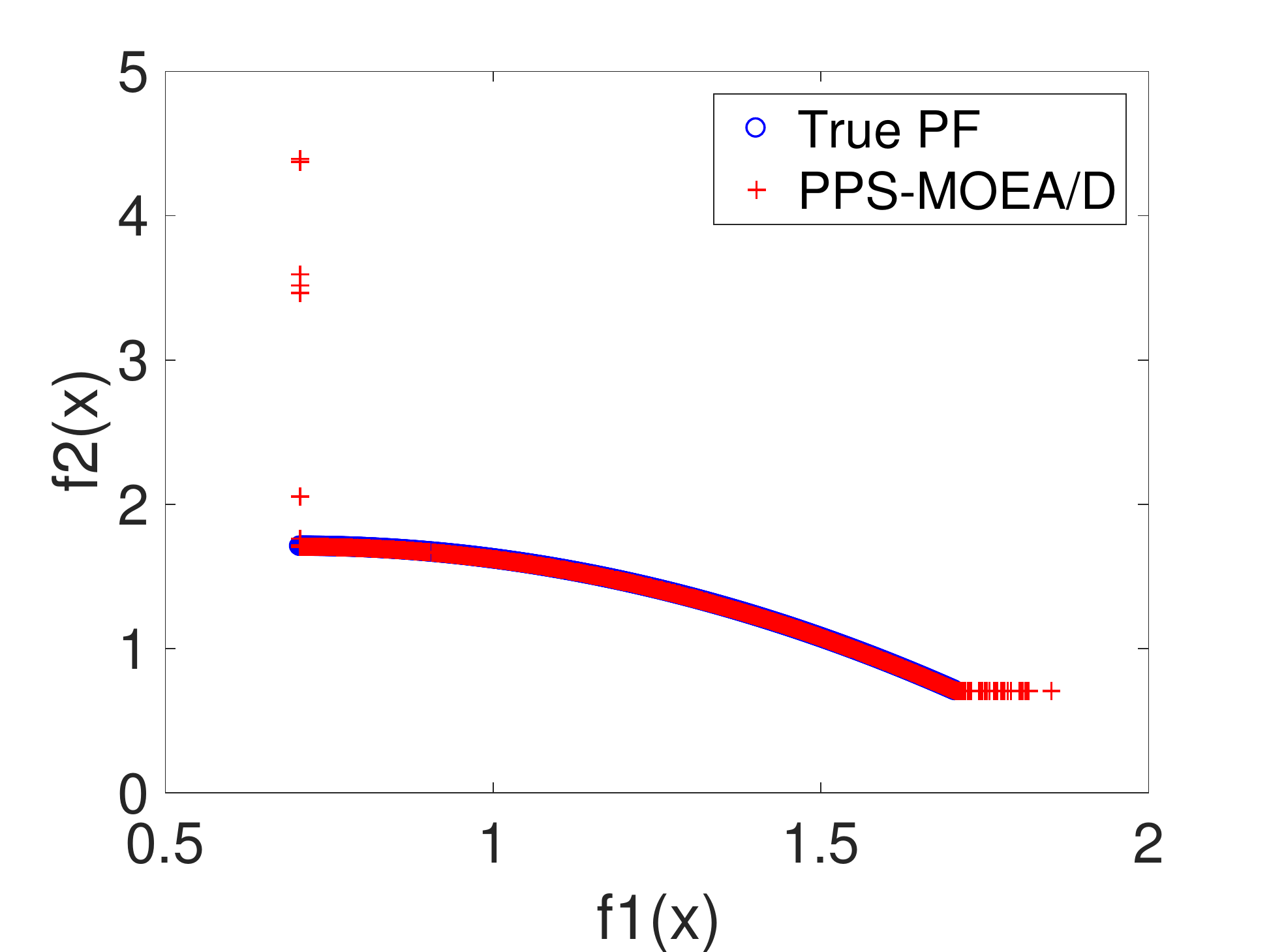}\\
\centering{\scriptsize{(a) PPS-MOEA/D}}
\end{minipage}
\hspace{0.5cm}
\begin{minipage}[t]{0.28\linewidth}
\includegraphics[width = 6.0cm]{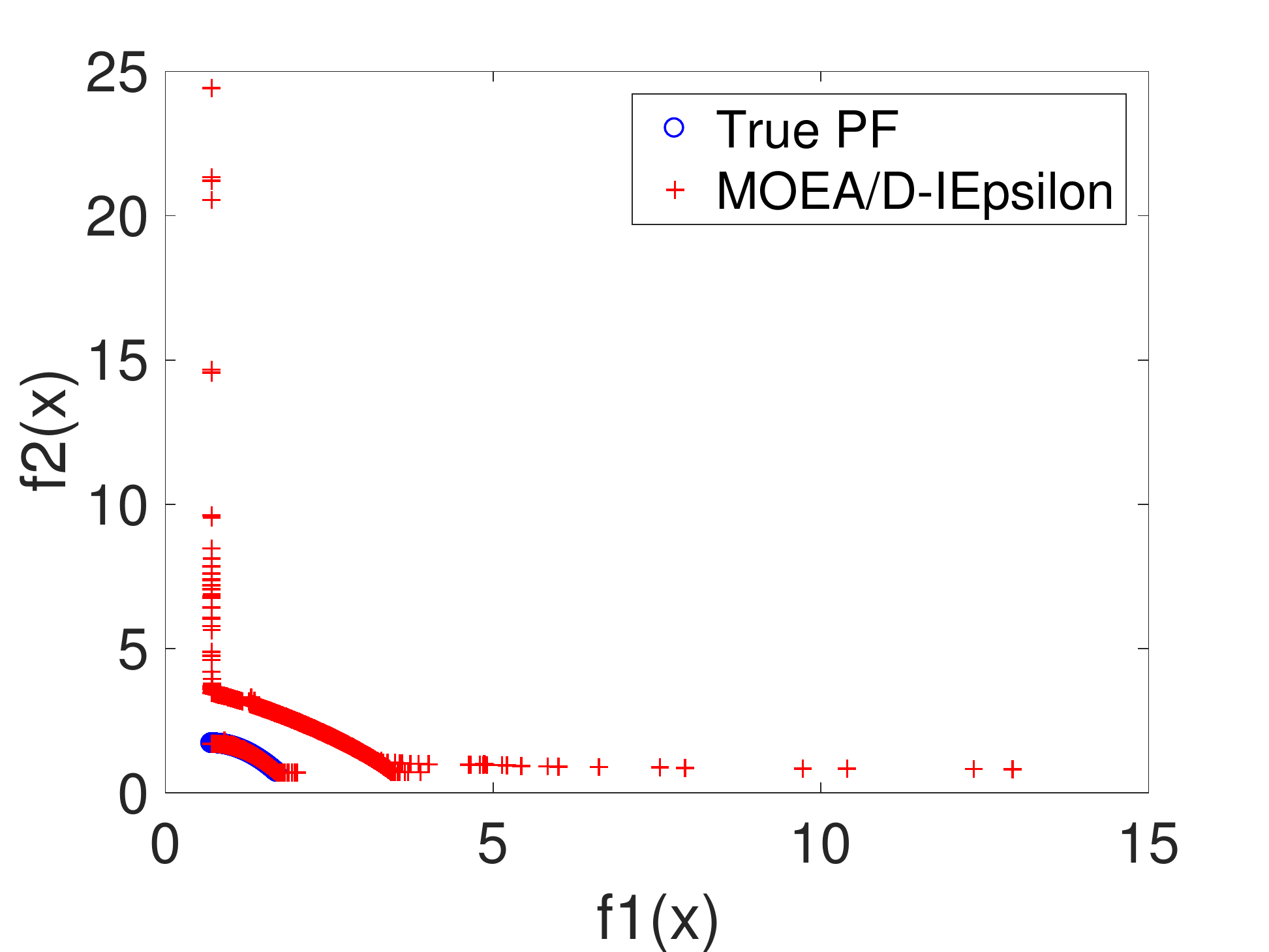}\\
\centering{\scriptsize{(b) MOEA/D-IEpsilon}}
\end{minipage}
\hspace{0.5cm}
\begin{minipage}[t]{0.28\linewidth}
\includegraphics[width = 6.0cm]{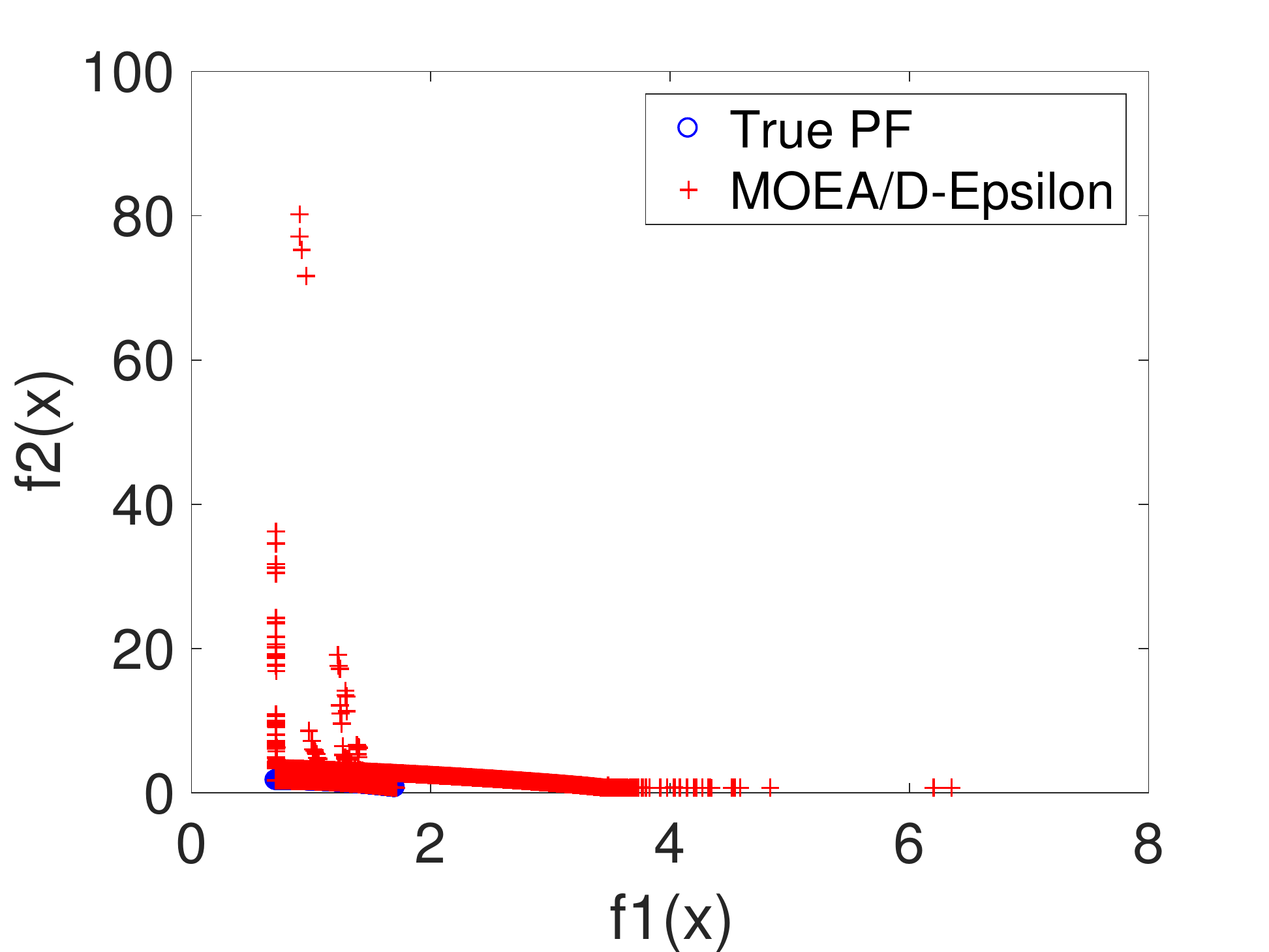}\\
\centering{\scriptsize{(c) MOEA/D-Epsilon}}
\end{minipage}
\end{tabular}

\vspace{0.2cm}
\begin{tabular}{cc}
\begin{minipage}[t]{0.28\linewidth}
\includegraphics[width = 6.0cm]{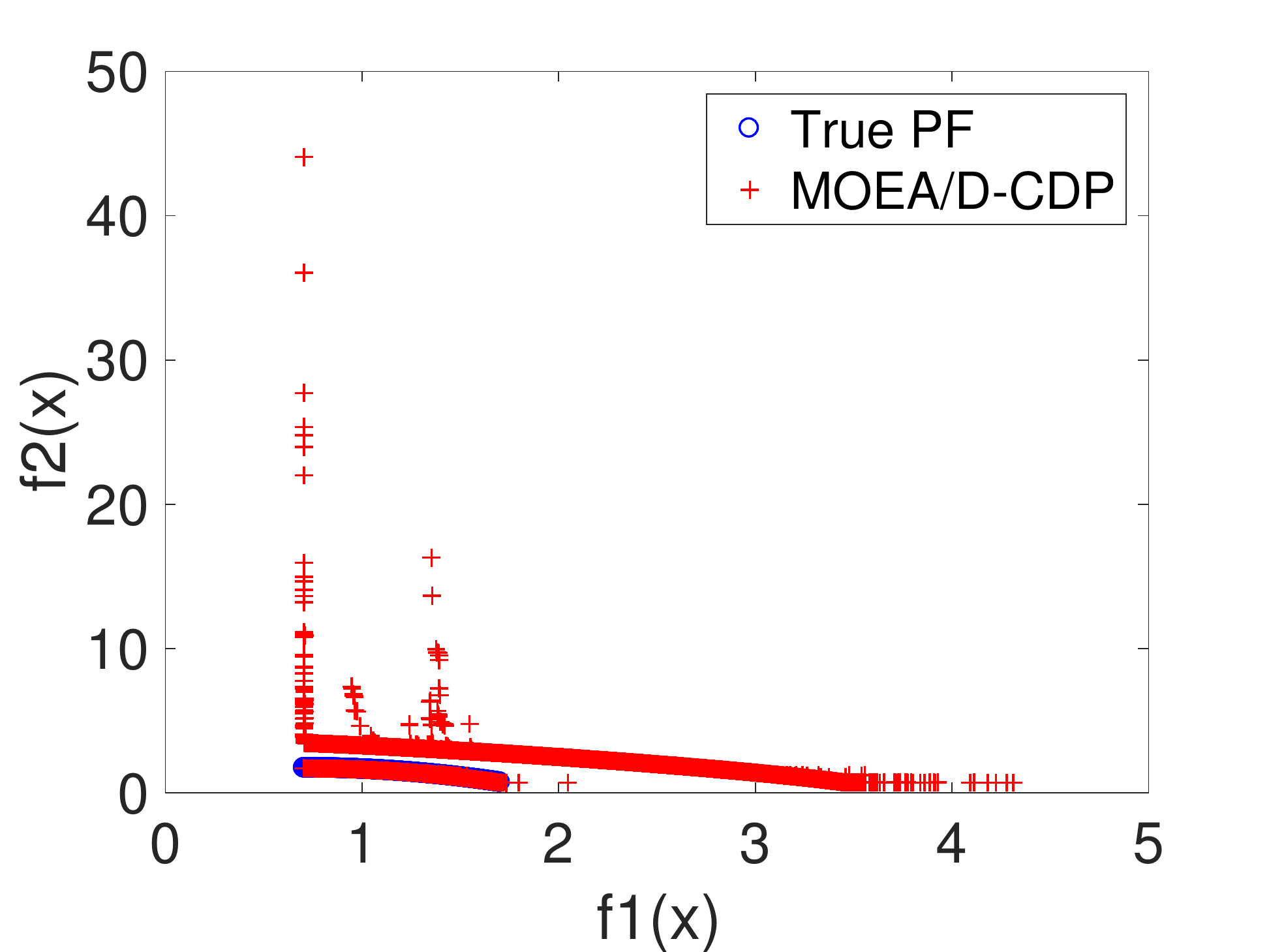}\\
\centering{\scriptsize{(d) MOEA/D-CDP}}
\end{minipage}
\hspace{0.5cm}
\begin{minipage}[t]{0.28\linewidth}
\includegraphics[width = 6.0cm]{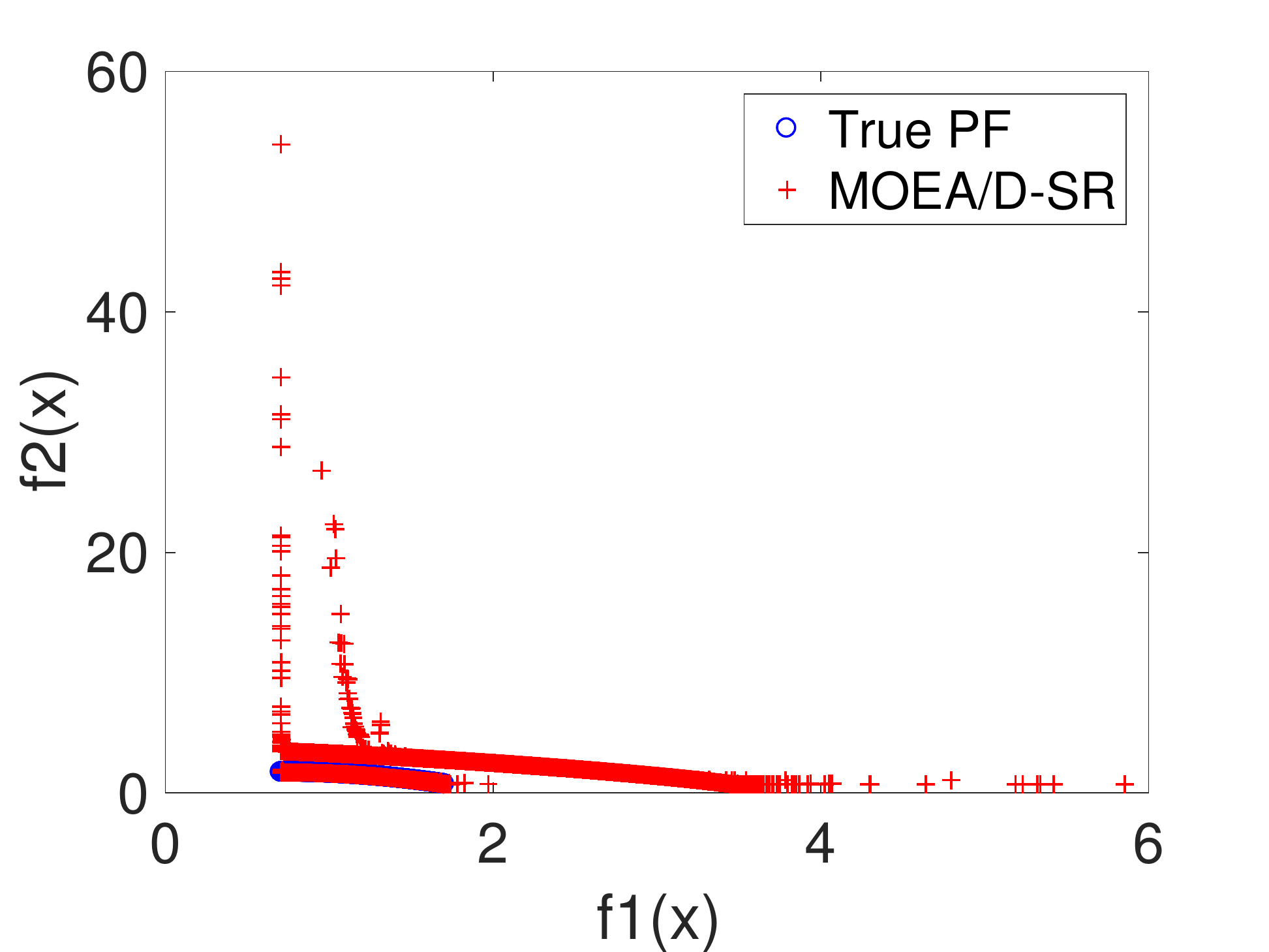}\\
\centering{\scriptsize{(e) MOEA/D-SR}}
\end{minipage}
\hspace{0.5cm}
\begin{minipage}[t]{0.28\linewidth}
\includegraphics[width = 6.0cm]{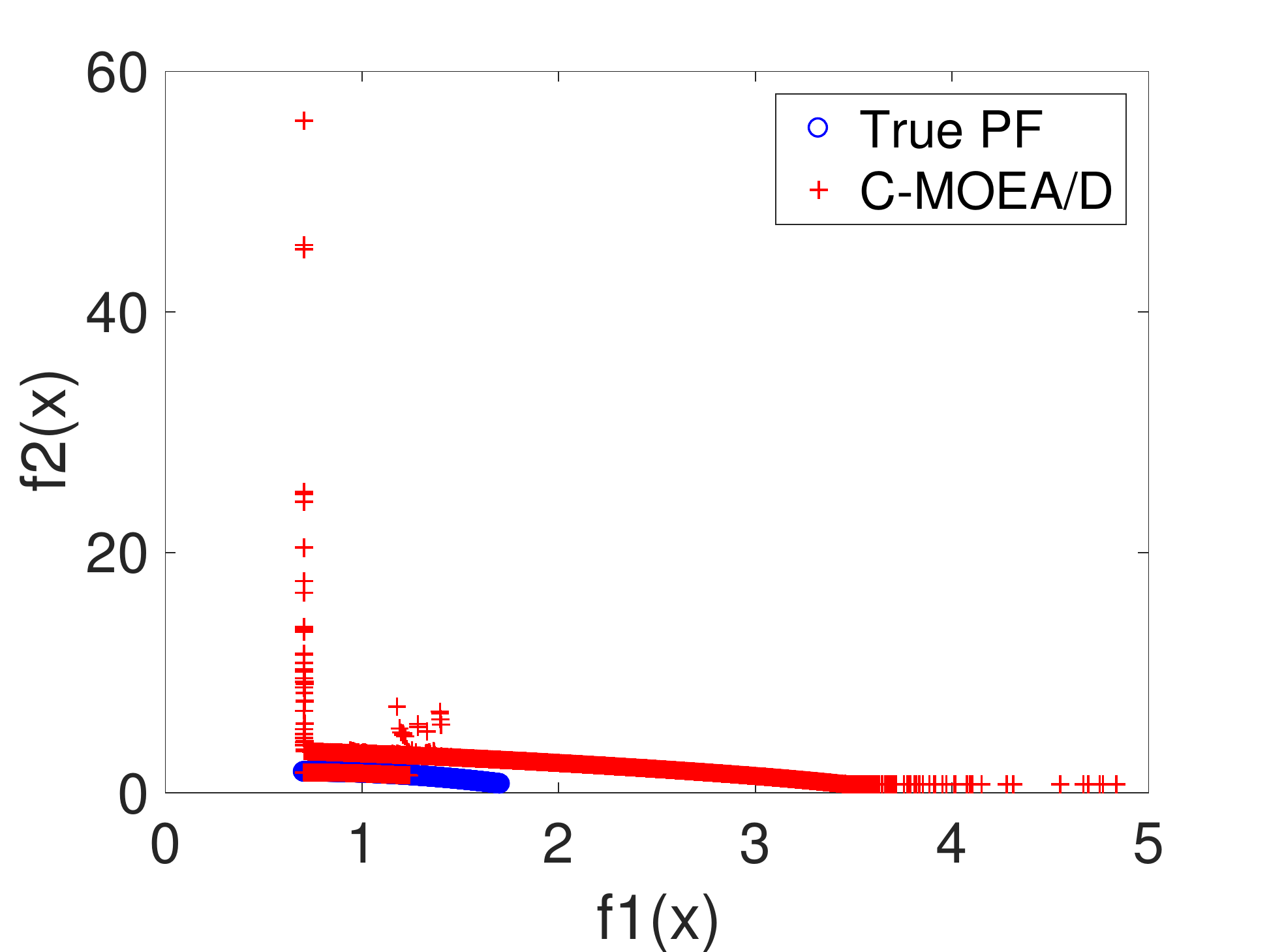}\\
\centering{\scriptsize{(f) C-MOEA/D}}
\end{minipage}
\end{tabular}

\caption{\label{fig:LIR-CMOP6-pops} The non-dominated solutions achieved by each algorithm on LIR-CMOP6 during the 30 independent runs are plotted in (a)-(f).}
\end{figure*}

\begin{figure*}
\begin{tabular}{cc}
\begin{minipage}[t]{0.28\linewidth}
\includegraphics[width = 6.0cm]{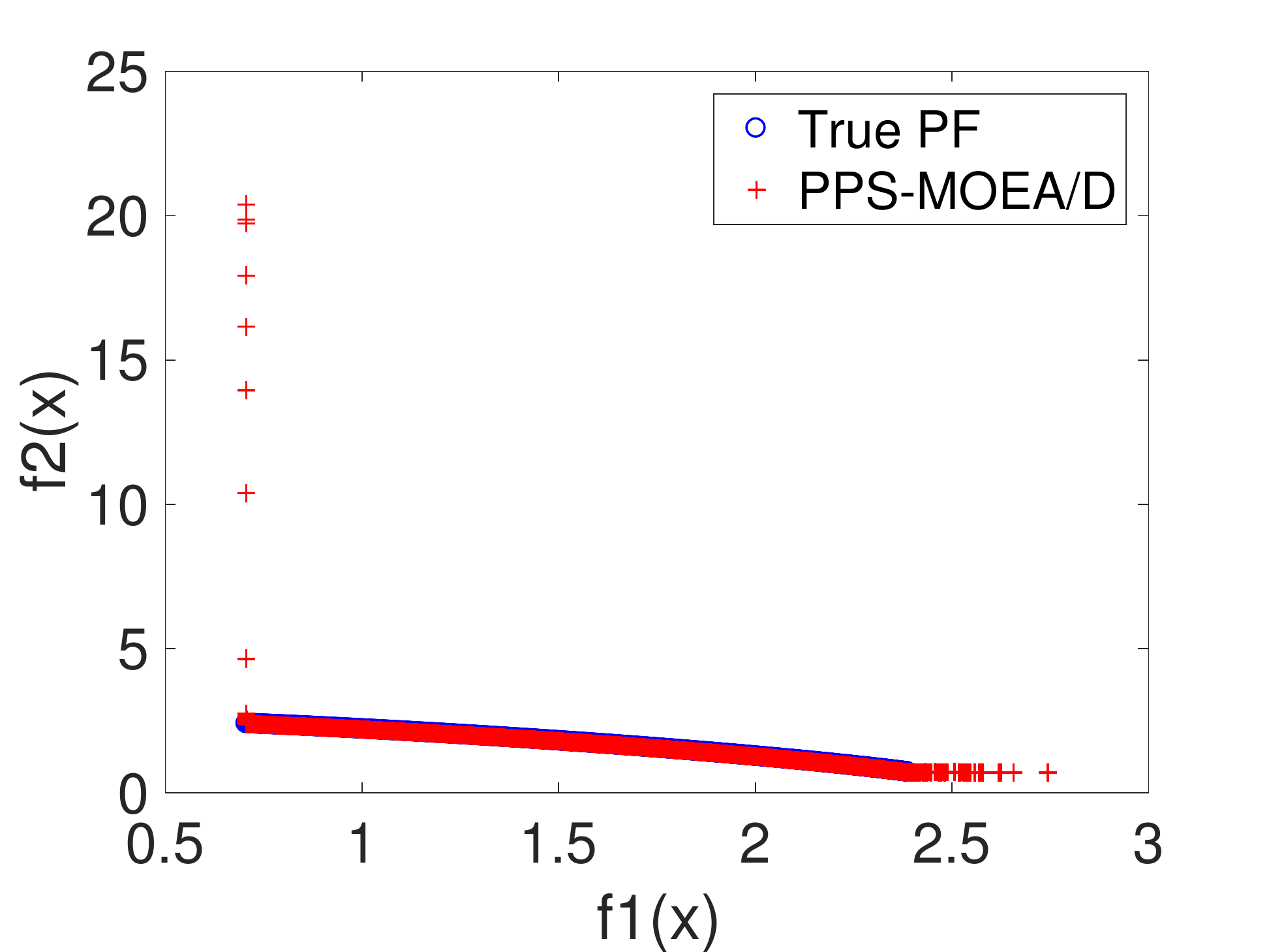}\\
\centering{\scriptsize{(a) PPS-MOEA/D}}
\end{minipage}
\hspace{0.5cm}
\begin{minipage}[t]{0.28\linewidth}
\includegraphics[width = 6.0cm]{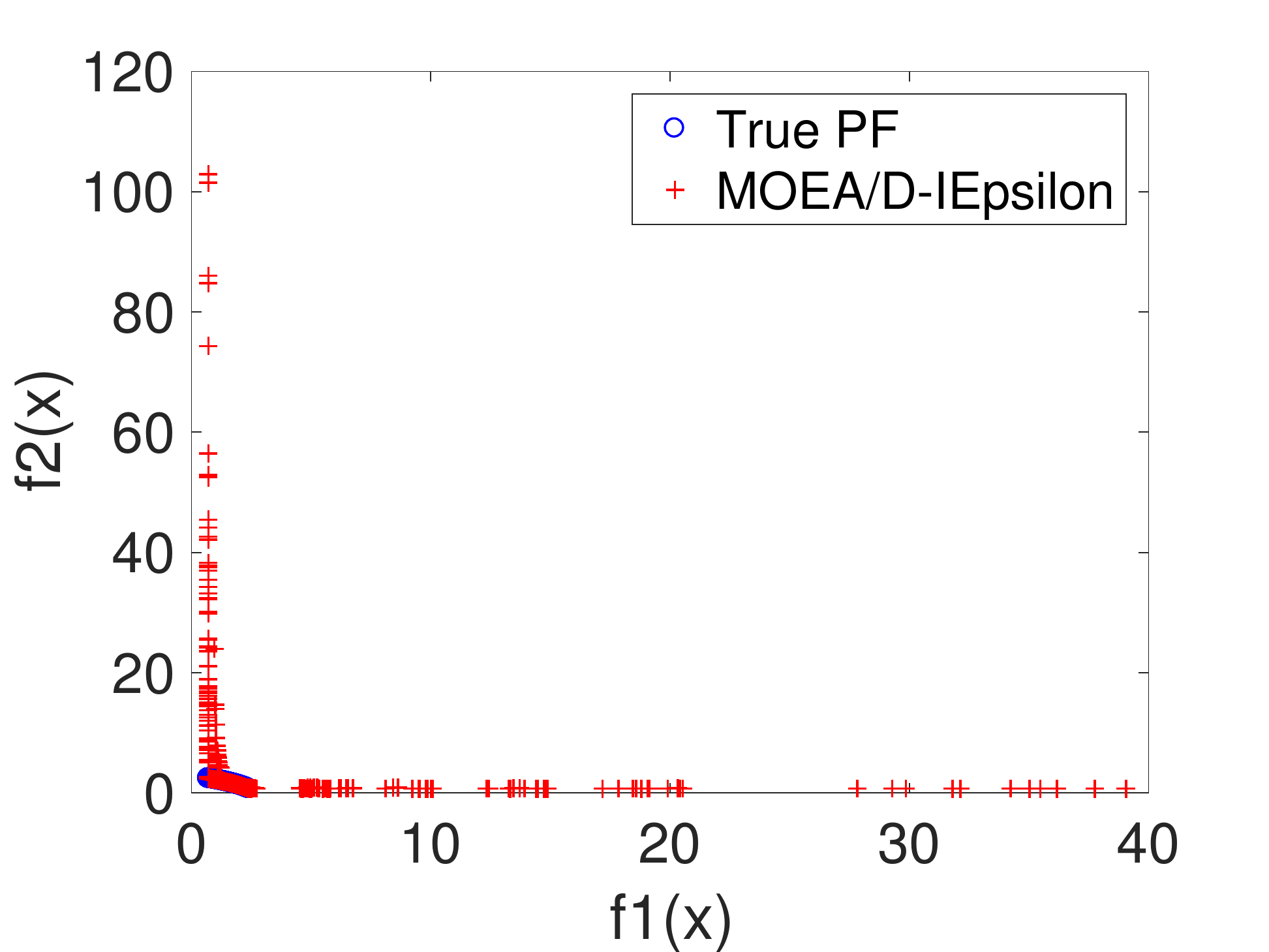}\\
\centering{\scriptsize{(b) MOEA/D-IEpsilon}}
\end{minipage}
\hspace{0.5cm}
\begin{minipage}[t]{0.28\linewidth}
\includegraphics[width = 6.0cm]{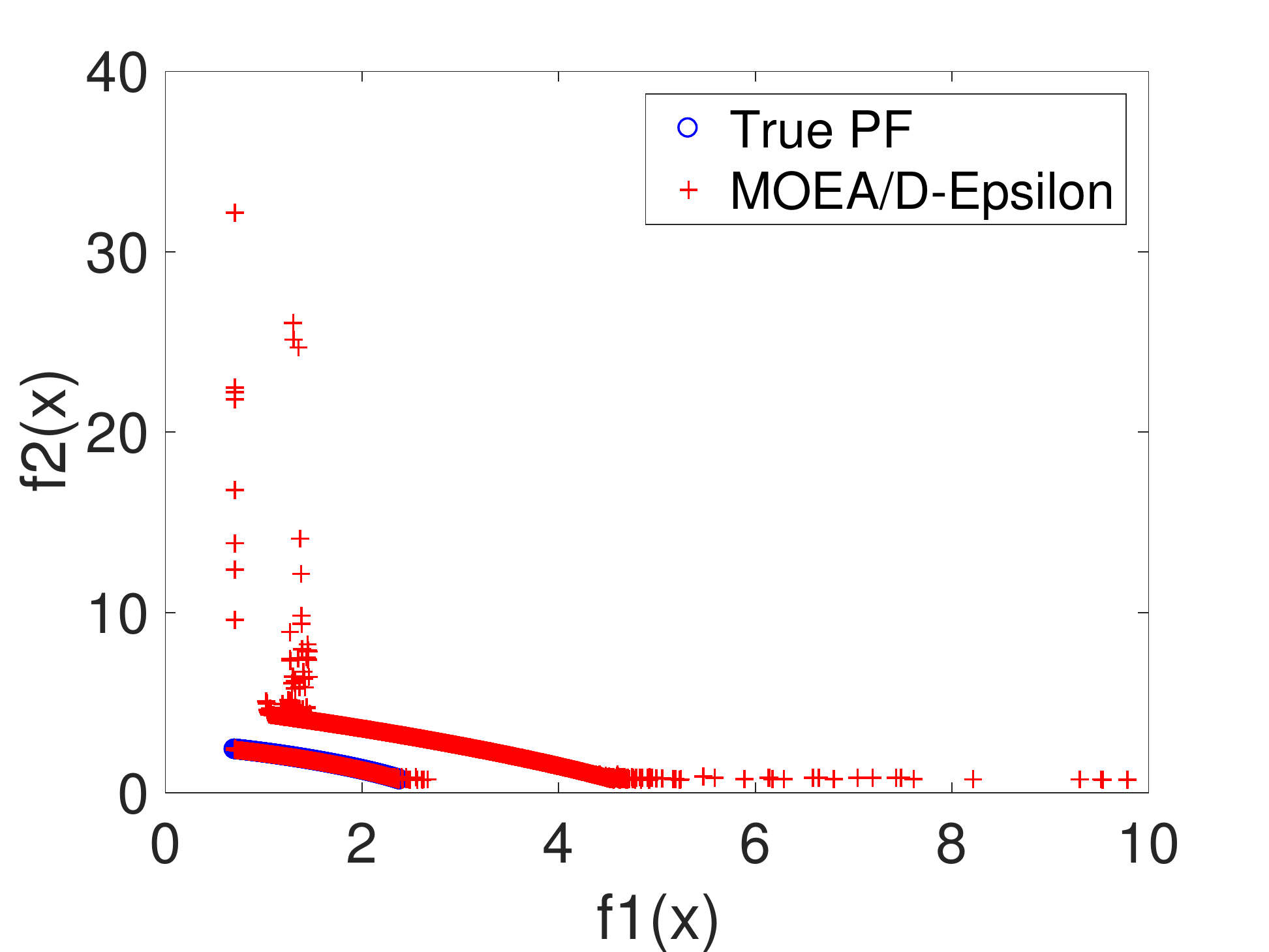}\\
\centering{\scriptsize{(c) MOEA/D-Epsilon}}
\end{minipage}
\end{tabular}

\vspace{0.2cm}
\begin{tabular}{cc}
\begin{minipage}[t]{0.28\linewidth}
\includegraphics[width = 6.0cm]{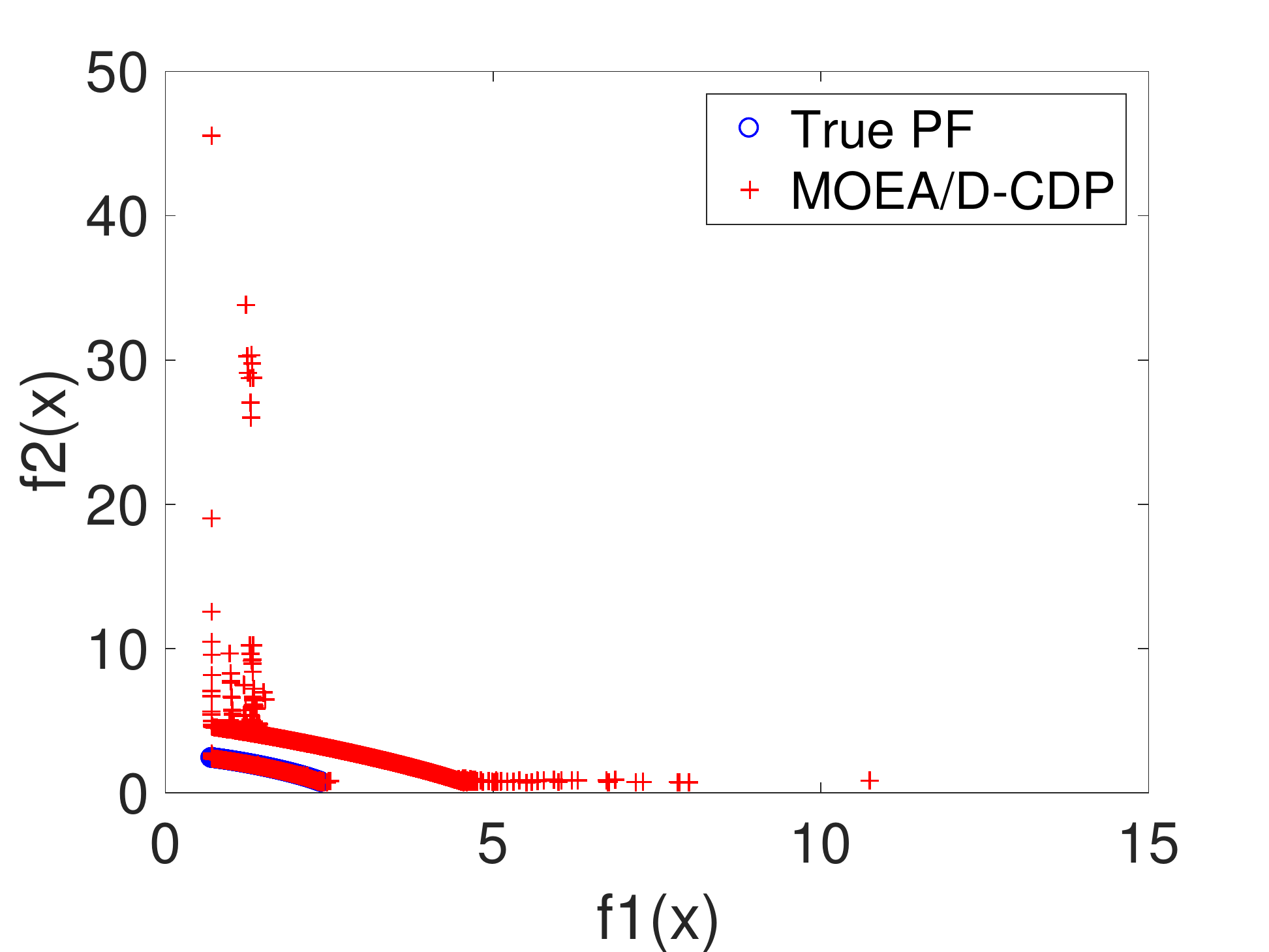}\\
\centering{\scriptsize{(d)} MOEA/D-CDP}
\end{minipage}
\hspace{0.5cm}
\begin{minipage}[t]{0.28\linewidth}
\includegraphics[width = 6.0cm]{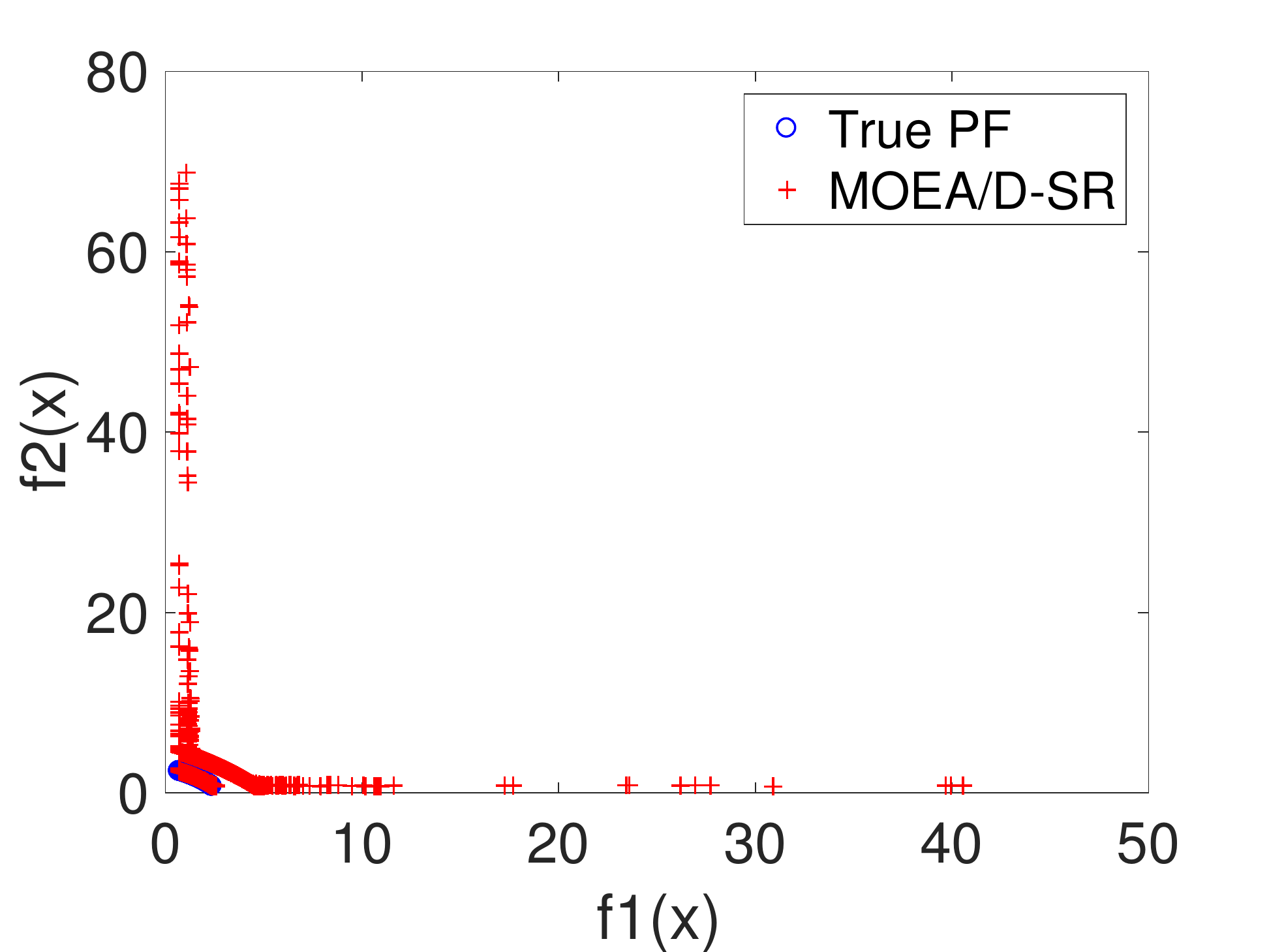}\\
\centering{\scriptsize{(e) MOEA/D-SR}}
\end{minipage}
\hspace{0.5cm}
\begin{minipage}[t]{0.28\linewidth}
\includegraphics[width = 6.0cm]{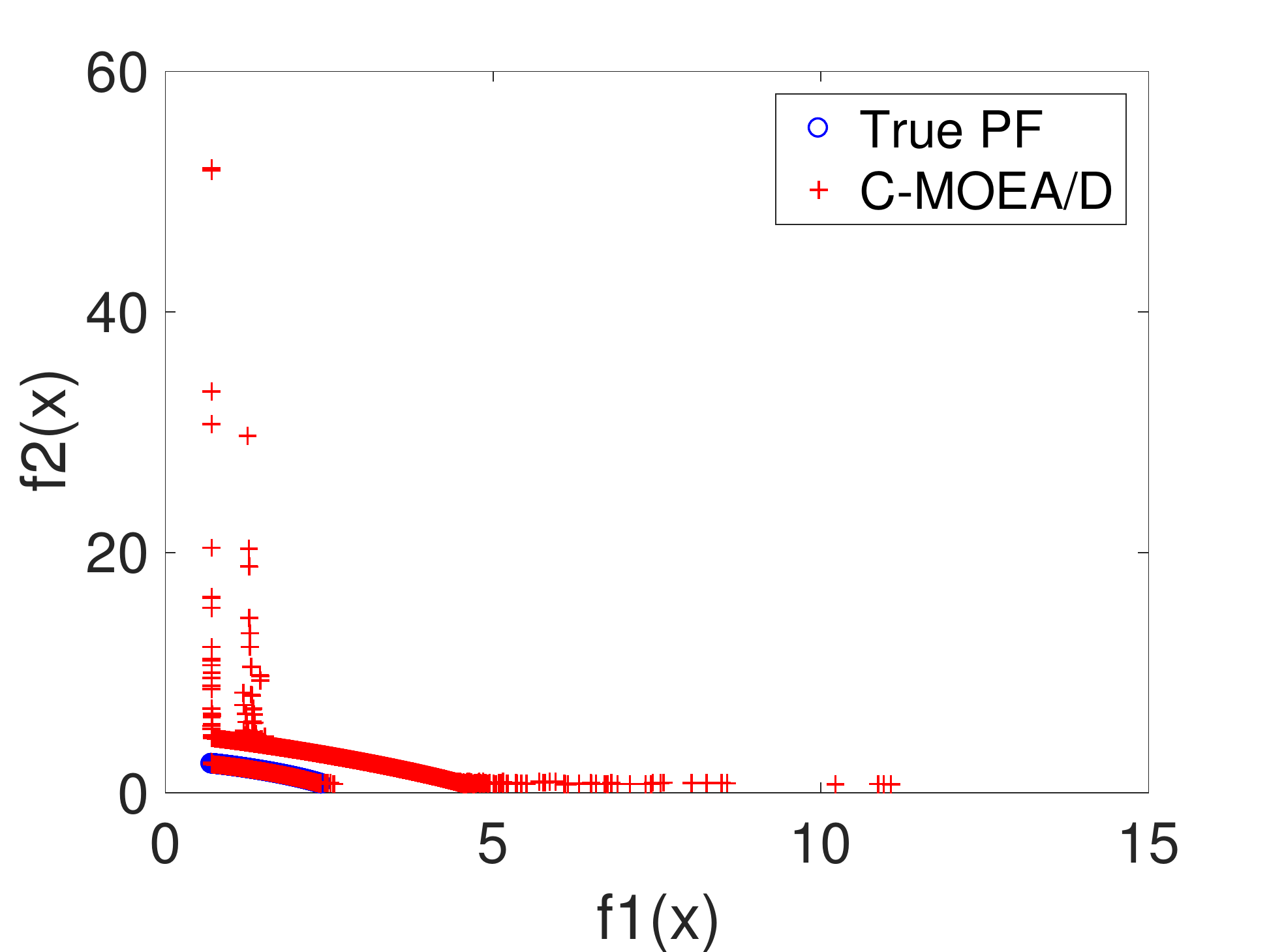}\\
\centering{\scriptsize{(f) C-MOEA/D}}
\end{minipage}
\end{tabular}

\caption{\label{fig:LIR-CMOP7-pops} The non-dominated solutions achieved by each algorithm on LIR-CMOP7 during the 30 independent runs are plotted in (a)-(f).}
\end{figure*}

\begin{figure*}
\begin{tabular}{cc}
\begin{minipage}[t]{0.28\linewidth}
\includegraphics[width = 6cm]{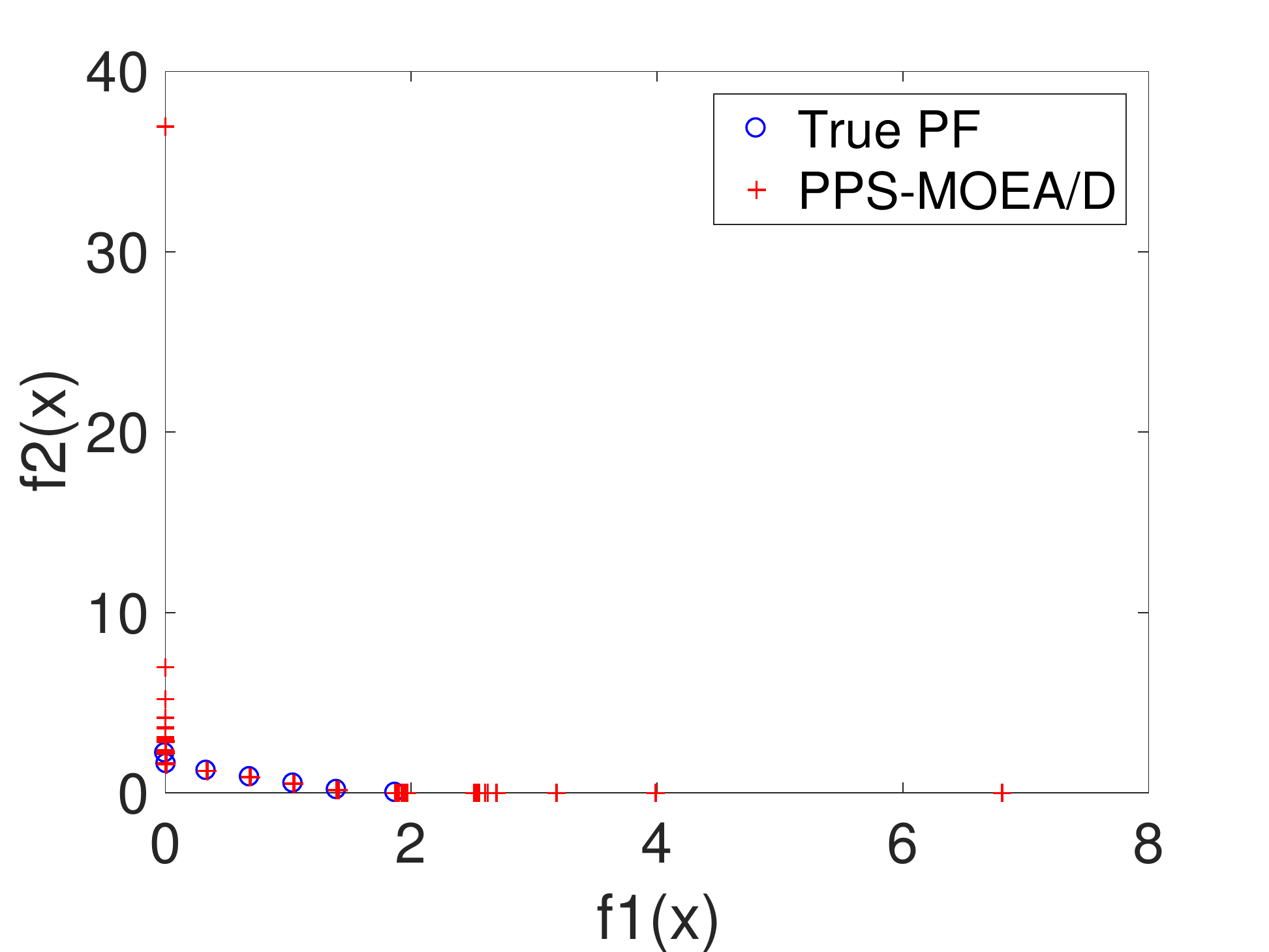}\\
\centering{\scriptsize{(a) PPS-MOEA/D}}
\end{minipage}
\hspace{0.5cm}
\begin{minipage}[t]{0.28\linewidth}
\includegraphics[width = 6cm]{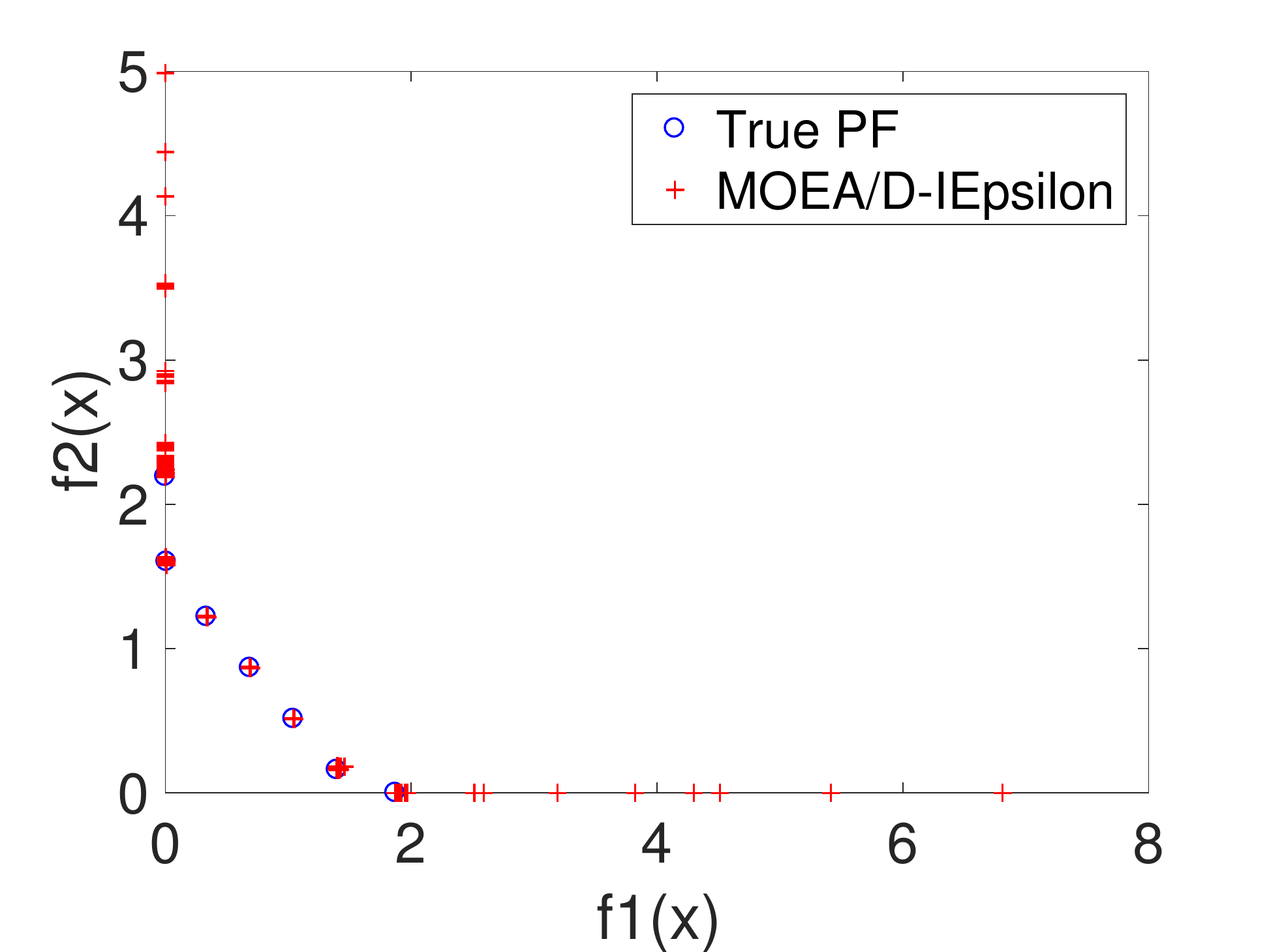}\\
\centering{\scriptsize{(b) MOEA/D-IEpsilon}}
\end{minipage}
\hspace{0.5cm}
\begin{minipage}[t]{0.28\linewidth}
\includegraphics[width = 6cm]{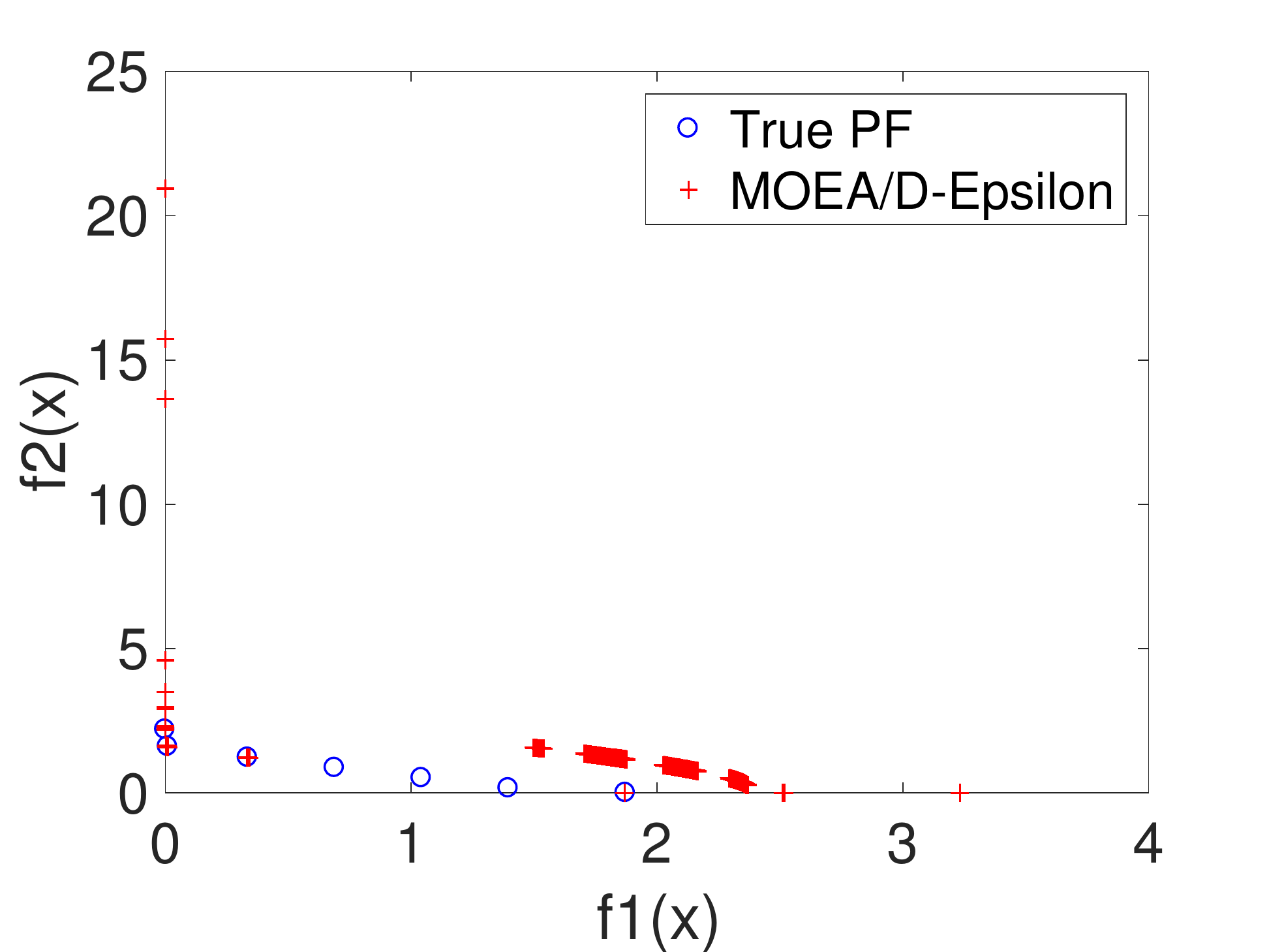}\\
\centering{\scriptsize{(c) MOEA/D-Epsilon}}
\end{minipage}
\end{tabular}

\vspace{0.2cm}
\begin{tabular}{cc}
\begin{minipage}[t]{0.28\linewidth}
\includegraphics[width = 6cm]{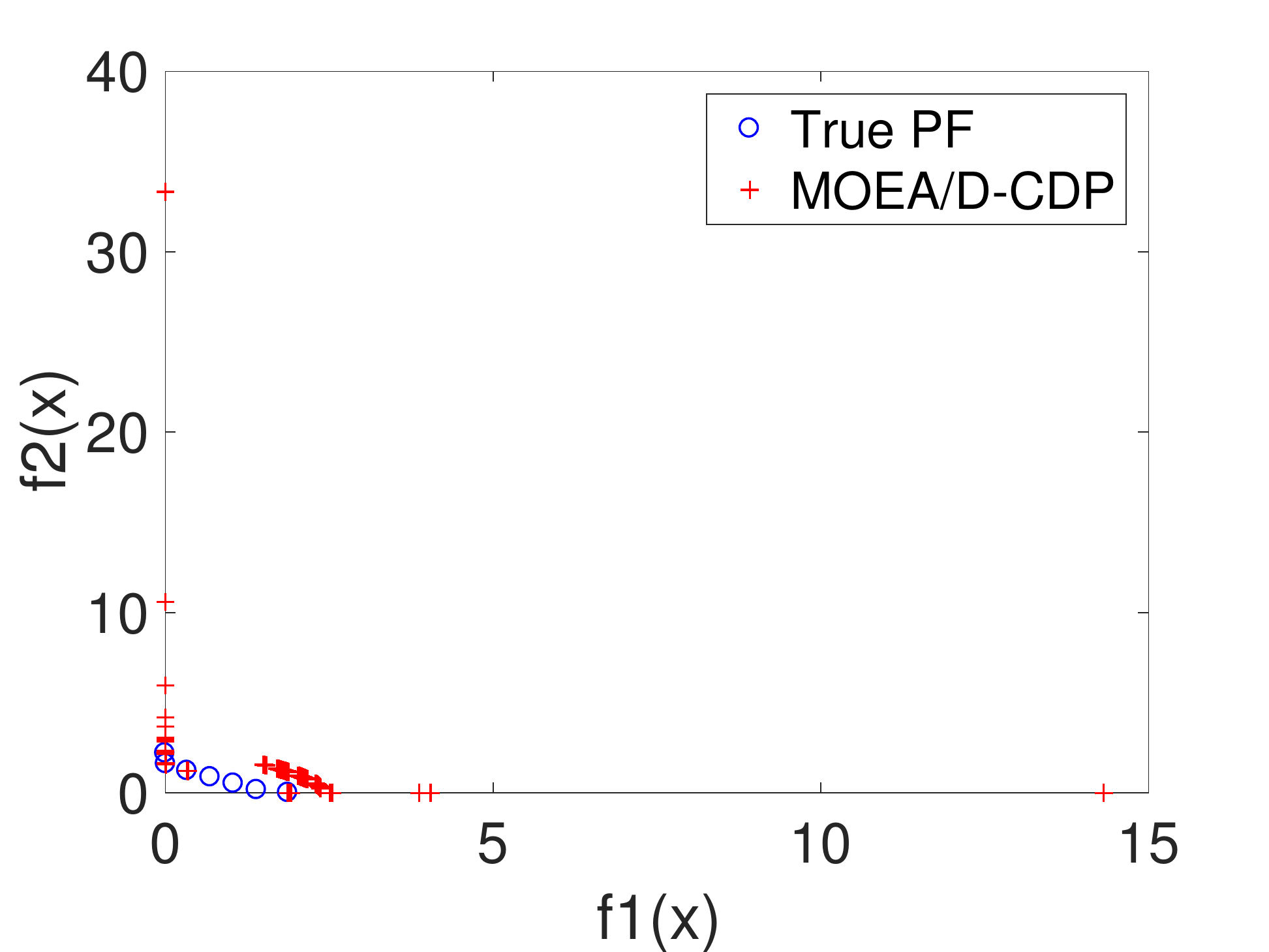}\\
\centering{\scriptsize{(d) MOEA/D-CDP}}
\end{minipage}
\hspace{0.5cm}
\begin{minipage}[t]{0.28\linewidth}
\includegraphics[width = 6cm]{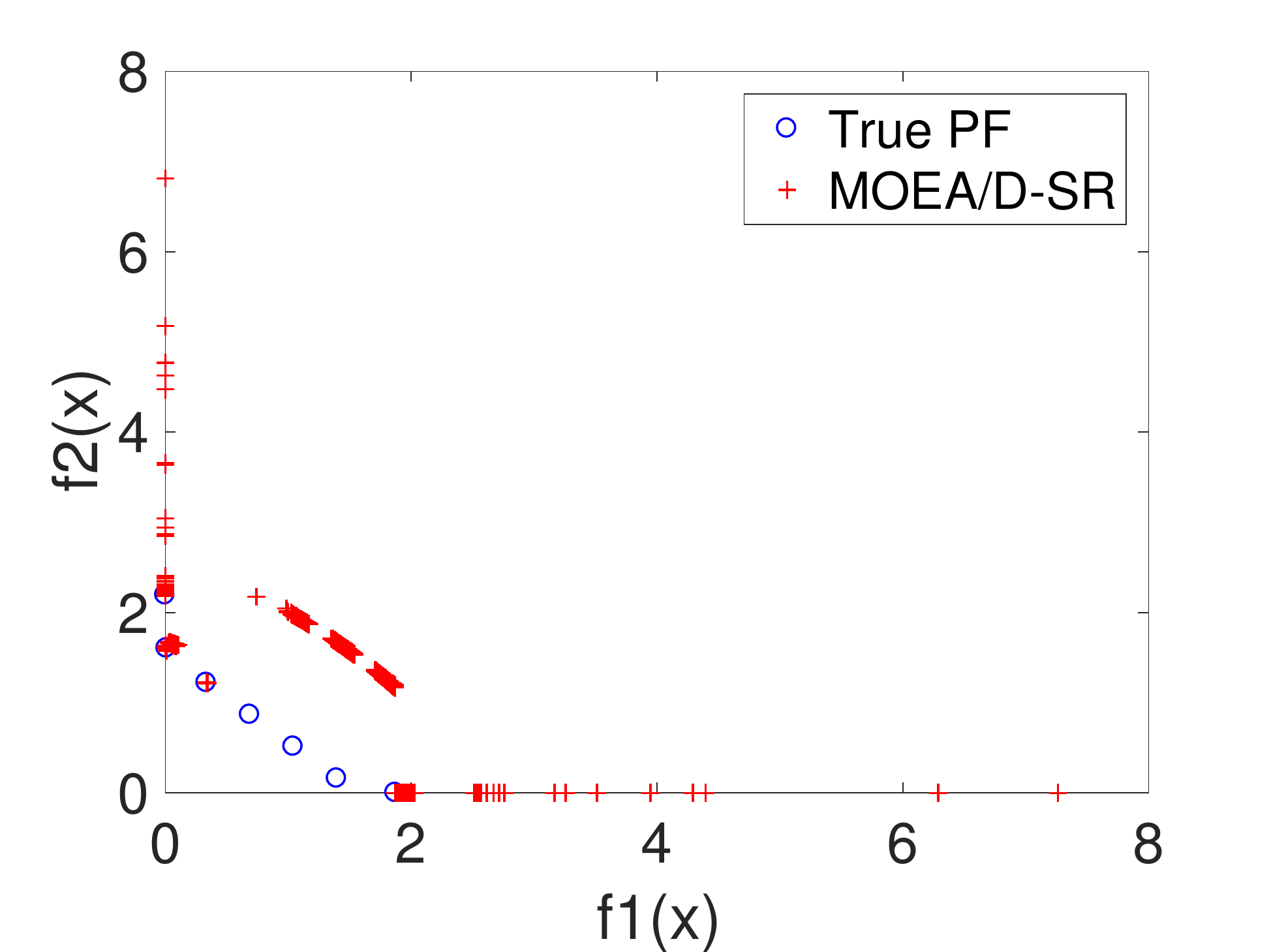}\\
\centering{\scriptsize{(e) MOEA/D-SR}}
\end{minipage}
\hspace{0.5cm}
\begin{minipage}[t]{0.28\linewidth}
\includegraphics[width = 6cm]{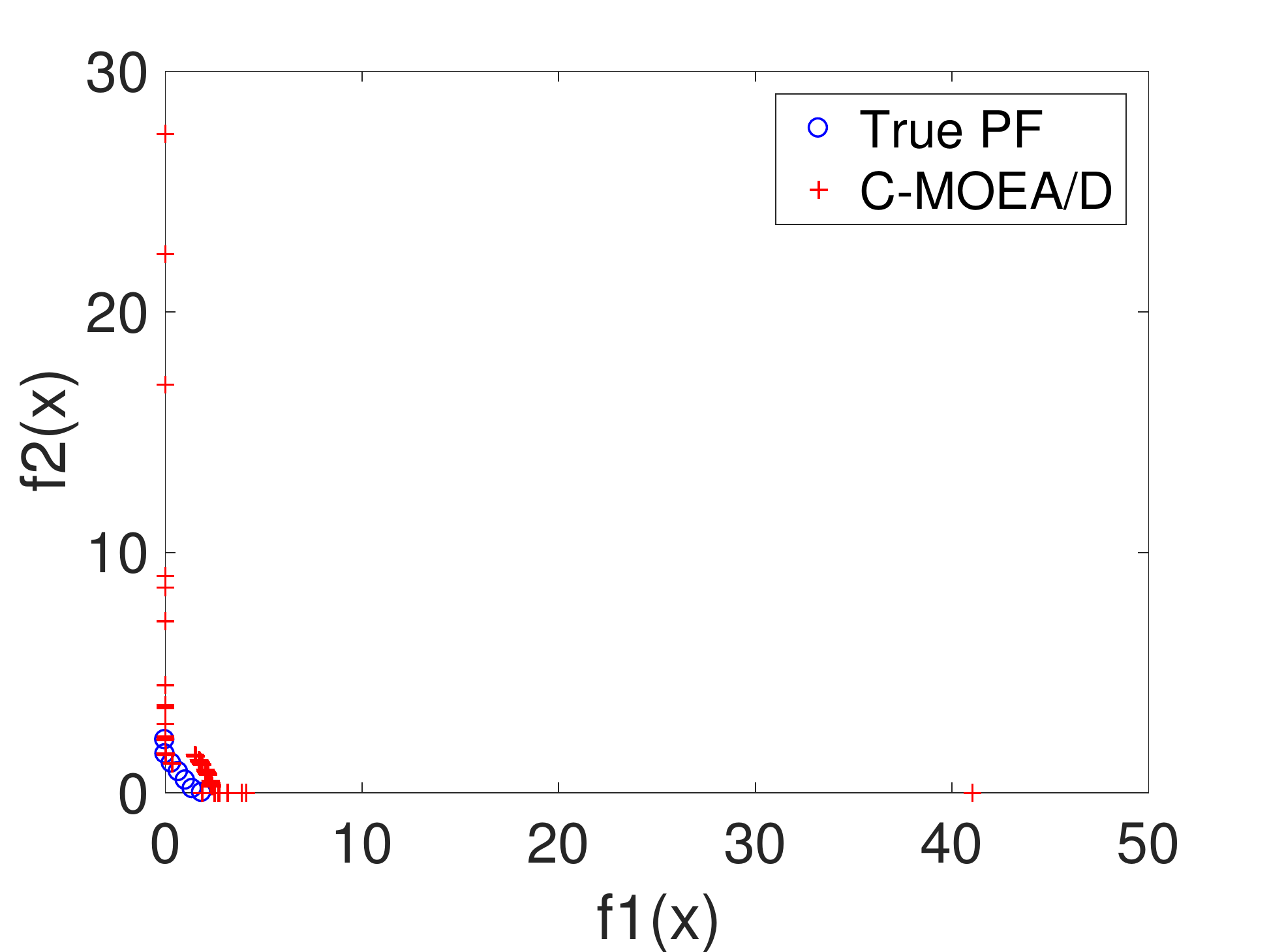}\\
\centering{\scriptsize{(f) C-MOEA/D}}
\end{minipage}
\end{tabular}

\caption{\label{fig:LIR-CMOP11-pops} The non-dominated solutions achieved by each algorithm on LIR-CMOP11 during the 30 independent runs are plotted in (a)-(f).}
\end{figure*}

The box plots of IGD values for the six tested CMOEAs on LIR-CMOP6, LIR-CMOP7 and LIR-CMOP11 are shown in Fig. \ref{fig:igd-box}. It is clear that the mean IGD value of PPS-MOEA/D is the smallest among the six tested CMOEAs. To further illustrate the advantages of the proposed PPS-MOEA/D, the populations achieved by each tested CMOEAs on LIR-CMOP6, LIR-CMOP7, and LIR-CMOP11 during the 30 independent runs are plotted in Fig. \ref{fig:LIR-CMOP6-pops}, Fig. \ref{fig:LIR-CMOP7-pops} and Fig. \ref{fig:LIR-CMOP11-pops}, respectively.

\begin{figure*}
\begin{tabular}{cc}
\begin{minipage}[t]{0.28\linewidth}
\includegraphics[width = 5.5cm]{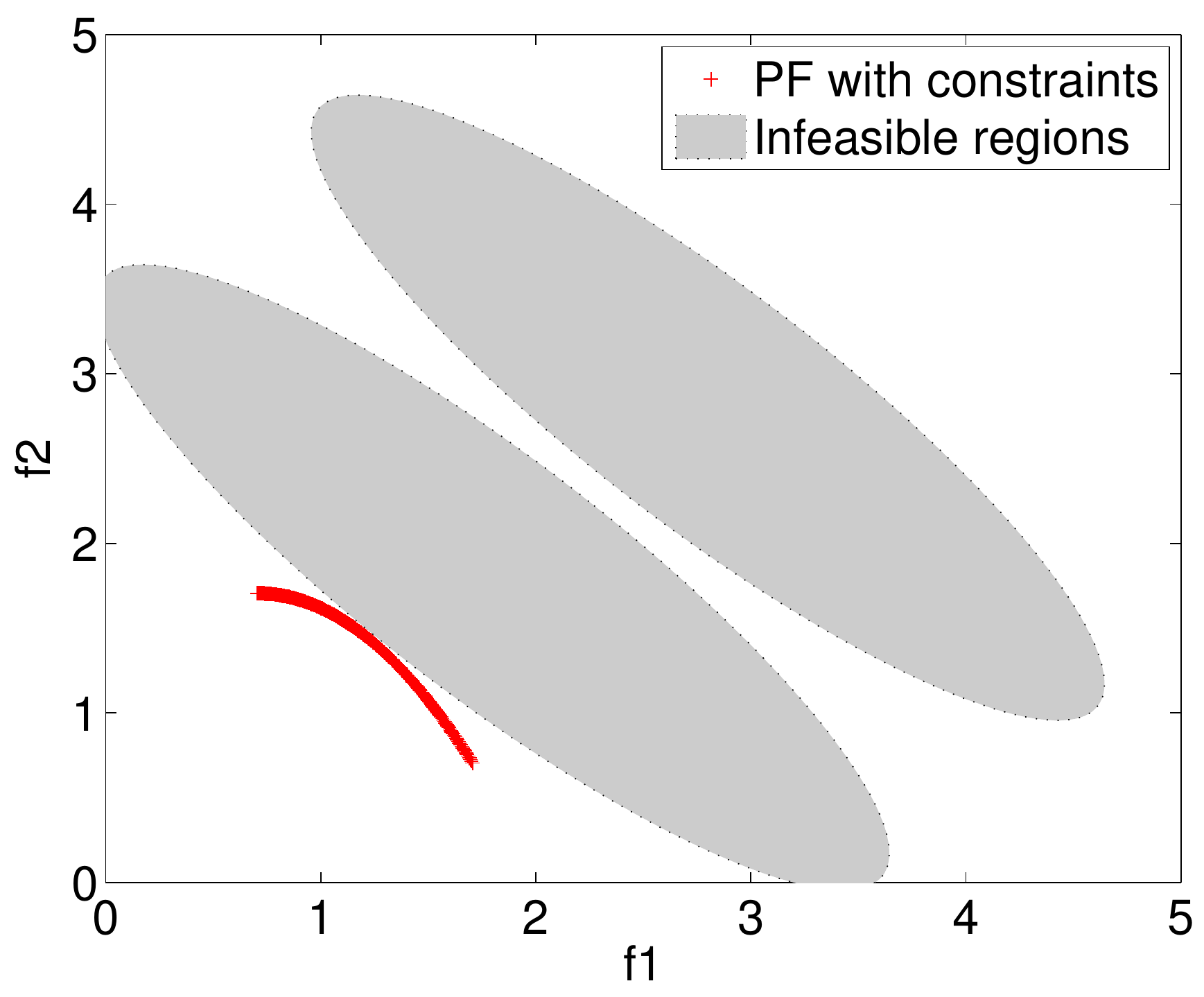}\\
\centering{\scriptsize{(a) LIR-CMOP6}}
\end{minipage}
\hspace{0.5cm}
\begin{minipage}[t]{0.28\linewidth}
\includegraphics[width = 5.5cm]{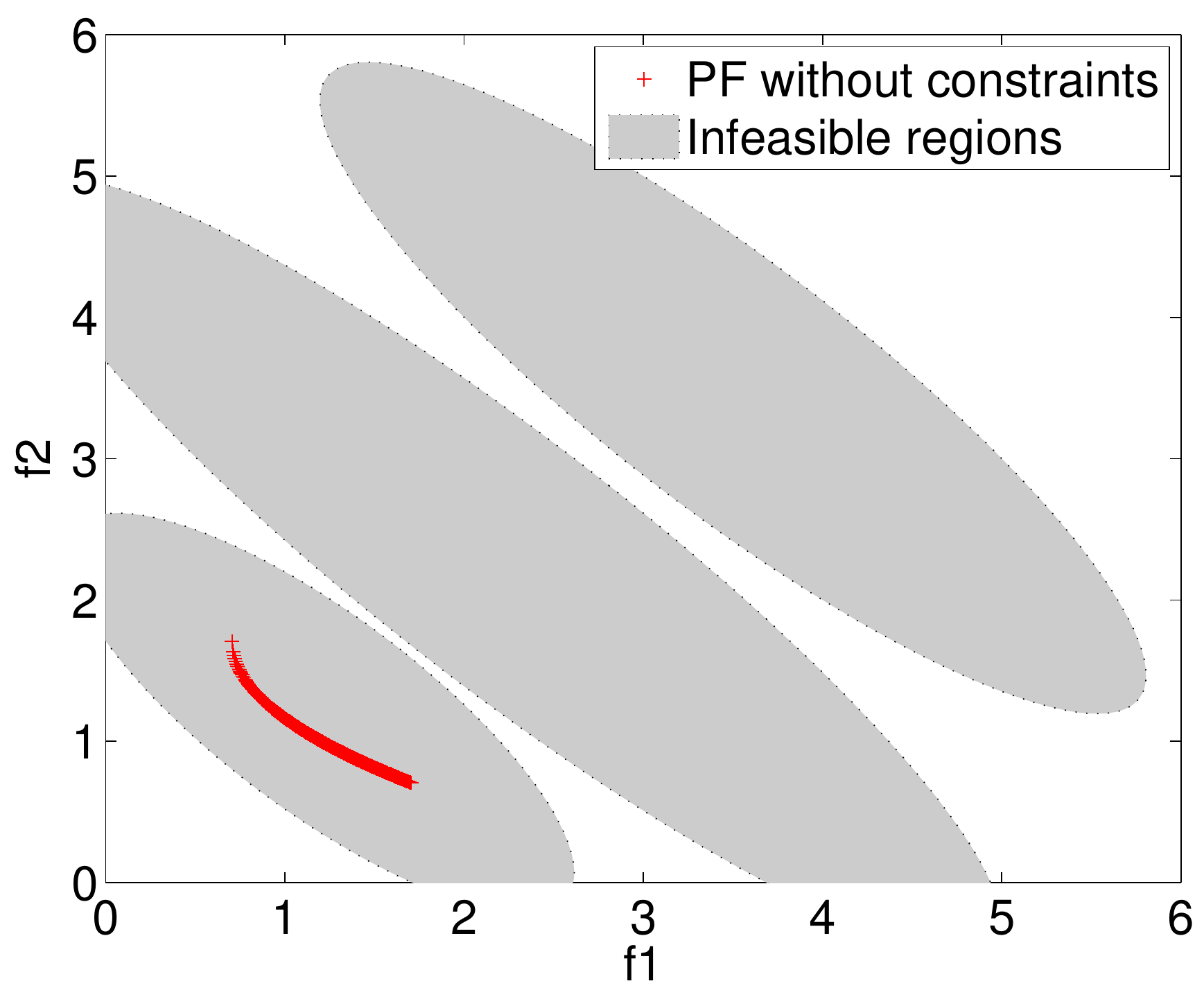}\\
\centering{\scriptsize{(b) LIR-CMOP7}}
\end{minipage}
\hspace{0.5cm}
\begin{minipage}[t]{0.28\linewidth}
\includegraphics[width = 5.5cm]{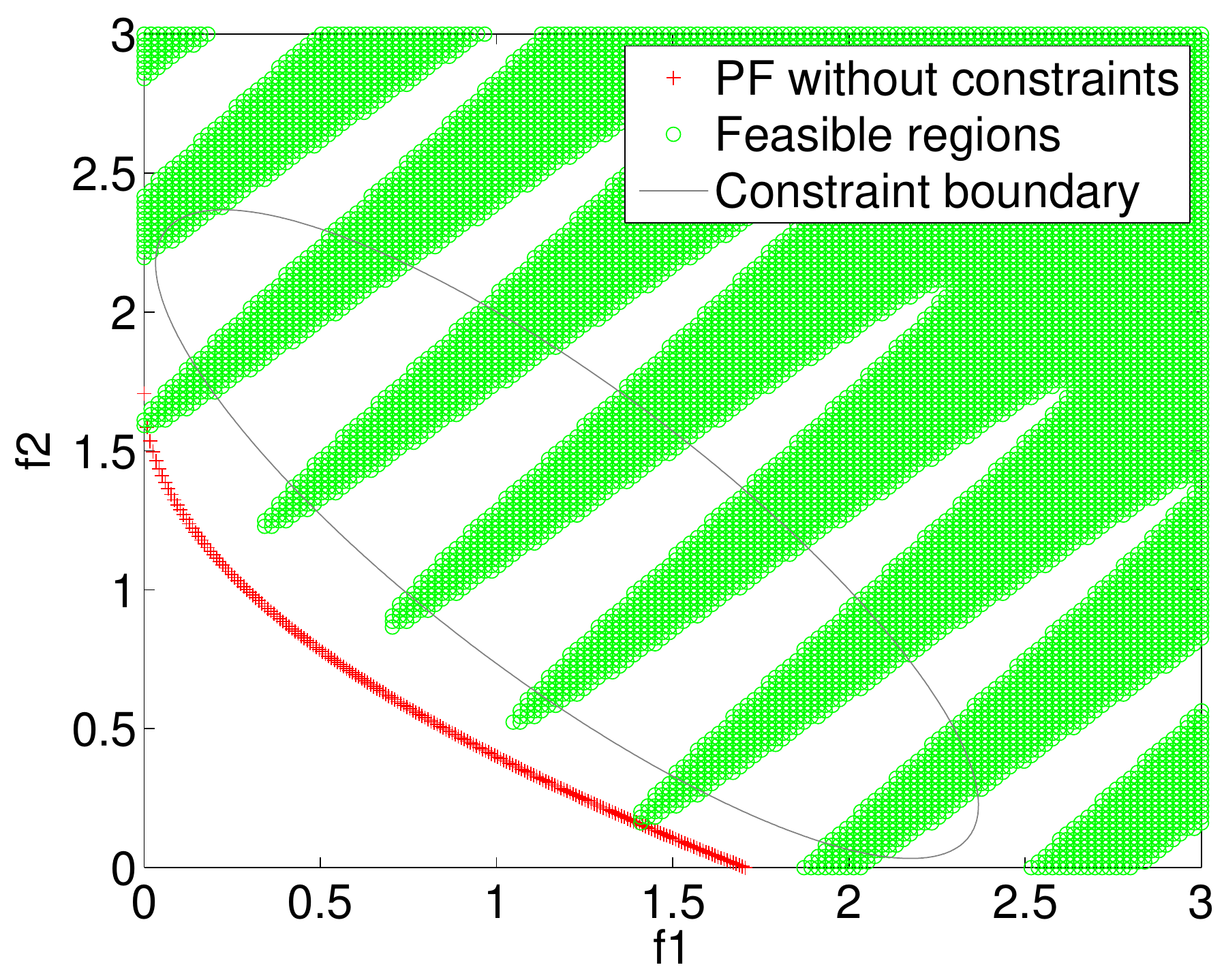}\\
\centering{\scriptsize{(c) LIR-CMOP11}}
\end{minipage}
\end{tabular}

\caption{\label{fig:pf} Illustrations of the feasible and infeasible regions of LIR-CMOP6, LIR-CMOP7 and LIR-CMOP11.}
\end{figure*}

For LIR-CMOP6, there are two large infeasible regions in front of the PF, and the unconstrained PF is the same as the constrained PF, as illustrated by Fig. \ref{fig:pf}(a). In Fig. \ref{fig:LIR-CMOP6-pops}, we can observe that PPS-MOEA/D can get across the large infeasible regions at each running time, while the other five CMOEAs are sometimes trapped into local optimal.

For LIR-CMOP7, there are three large infeasible regions, and the unconstrained PF is covered by infeasible regions and becomes no more feasible as illustrated by Fig. \ref{fig:pf}(b). In Fig. \ref{fig:LIR-CMOP7-pops}, we can observe that PPS-MOEA/D can converge to the real PF at each running time, while the other five CMOEAs can not always converge to the real PF.

For LIR-CMOP11, infeasible regions make the original unconstrained PF partially feasible, as illustrated by Fig. \ref{fig:pf}(c). The PF of LIR-CMOP11 is disconnected and has seven Pareto optimal solutions. PPS-MOEA/D can find the seven discrete Pareto optimal solutions at each running time, while the other five CMOEAs can not find all the seven Pareto optimal solutions sometimes, as shown in Fig. \ref{fig:LIR-CMOP11-pops}.  In Fig. \ref{fig:LIR-CMOP11-pops}(b), it seems that MOEA/D-IEpsilon has found all the Pareto optimal solutions. However, MOEA/D-IEpsilon can not find all the Pareto optimal solutions more than fifteen out of 30 runs. For MOEA/D-Epsilon, MOEA/D-CDP, MOEA/D-SR and C-MOEA/D, some Pareto optimal solutions are not found during the thirty independent runs, as illustrated by \ref{fig:LIR-CMOP11-pops}(c)-(f). 


\begin{table*}[htbp]
  \centering
  \caption{IGD results of PPS and the other five CMOEAs on LIR-CMOP1-14.}
    \begin{tabular}{c|c|c|c|c|c|c|c}
    \toprule
    \multicolumn{2}{c|}{Test Instances} & PPS-MOEA/D   & MOEA/D-IEpsilon & MOEA/D-Epsilon & MOEA/D-CDP & MOEA/D-SR & C-MOEA/D \\
    \hline
    \multirow{2}[0]{*}{LIR-CMOP1} & mean  & \textbf{6.413E-03} & 7.972E-03$\dag$ & 5.745E-02$\dag$ & 1.107E-01$\dag$ & 1.811E-02$\dag$ & 1.256E-01$\dag$ \\
          & std   & 1.938E-03 & 3.551E-03 & 2.887E-02 & 5.039E-02 & 1.665E-02 & 7.028E-02 \\
    \hline
    \multirow{2}[0]{*}{LIR-CMOP2} & mean  & \textbf{4.673E-03} & 5.234E-03$\dag$ & 5.395E-02$\dag$ & 1.434E-01$\dag$ & 9.634E-03$\dag$ & 1.397E-01$\dag$ \\
          & std   & 7.844E-04 & 1.009E-03 & 2.128E-02 & 6.055E-02 & 7.232E-03 & 5.436E-02 \\
    \hline
    \multirow{2}[0]{*}{LIR-CMOP3} & mean  & \textbf{8.6.0E-03} & 1.131E-02 & 8.806E-02$\dag$ & 2.608E-01$\dag$ & 1.775E-01$\dag$ & 2.805E-01$\dag$ \\
          & std   & 5.184E-03 & 6.422E-03 & 4.356E-02 & 4.335E-02 & 7.196E-02 & 4.214E-02 \\
    \hline
    \multirow{2}[0]{*}{LIR-CMOP4} & mean  & \textbf{4.677E-03} & 4.848E-03 & 6.506E-02$\dag$ & 2.527E-01$\dag$ & 1.949E-01$\dag$ & 2.590E-01$\dag$ \\
          & std   & 1.116E-03 & 2.051E-03 & 3.011E-02 & 4.340E-02 & 6.401E-02 & 3.513E-02 \\
    \hline
    \multirow{2}[0]{*}{LIR-CMOP5} & mean  & \textbf{1.837E-03} & 2.132E-03$\dag$ & 1.150E+00$\dag$ & 1.045E+00$\dag$ & 1.040E+00$\dag$ & 1.103E+00$\dag$ \\
          & std   & 9.263E-05 & 3.791E-04 & 1.979E-01 & 3.632E-01 & 3.655E-01 & 2.994E-01 \\
    \hline
    \multirow{2}[0]{*}{LIR-CMOP6} & mean  & \textbf{2.490E-03} & 2.334E-01$\dag$ & 1.269E+00$\dag$ & 1.090E+00$\dag$ & 9.427E-01$\dag$ & 1.307E+00$\dag$ \\
          & std   & 3.399E-04 & 5.062E-01 & 2.947E-01 & 5.198E-01 & 5.899E-01 & 2.080E-01 \\
    \hline
    \multirow{2}[0]{*}{LIR-CMOP7} & mean  & \textbf{2.797E-03} & 3.735E-02$\dag$ & 1.509E+00$\dag$ & 1.460E+00$\dag$ & 1.079E+00$\dag$ & 1.564E+00$\dag$ \\
          & std   & 9.854E-05 & 5.414E-02 & 5.094E-01 & 6.082E-01 & 7.581E-01 & 4.235E-01 \\
    \hline
    \multirow{2}[0]{*}{LIR-CMOP8} & mean  & \textbf{2.778E-03} & 2.749E-02$\dag$ & 1.620E+00$\dag$ & 1.375E+00$\dag$ & 1.012E+00$\dag$ & 1.579E+00$\dag$ \\
          & std   & 7.558E-05 & 5.917E-02 & 3.054E-01 & 6.149E-01 & 7.244E-01 & 3.706E-01 \\
    \hline
    \multirow{2}[0]{*}{LIR-CMOP9} & mean  & 9.940E-02 & \textbf{4.977E-03}$\ddag$ & 4.902E-01$\dag$ & 4.810E-01$\dag$ & 4.854E-01$\dag$ & 4.810E-01$\dag$ \\
          & std   & 1.519E-01 & 1.367E-02 & 4.221E-02 & 5.244E-02 & 4.775E-02 & 5.243E-02 \\
    \hline
    \multirow{2}[0]{*}{LIR-CMOP10} & mean  & \textbf{2.108E-03} & 2.111E-03 & 2.132E-01$\dag$ & 2.159E-01$\dag$ & 1.925E-01$\dag$ & 2.130E-01$\dag$ \\
          & std   & 7.754E-05 & 7.111E-05 & 5.315E-02 & 6.810E-02 & 6.812E-02 & 4.634E-02 \\
    \hline
    \multirow{2}[0]{*}{LIR-CMOP11} & mean  & \textbf{2.832E-03} & 5.809E-02$\dag$ & 3.468E-01$\dag$ & 3.418E-01$\dag$ & 3.157E-01$\dag$ & 3.806E-01$\dag$ \\
          & std   & 1.359E-03 & 5.793E-02 & 9.285E-02 & 9.216E-02 & 7.491E-02 & 8.949E-02 \\
    \hline
    \multirow{2}[0]{*}{LIR-CMOP12} & mean  & \textbf{2.704E-02} & 3.358E-02 & 2.520E-01$\dag$ & 2.690E-01$\dag$ & 2.064E-01$\dag$ & 2.6.0E-01$\dag$ \\
          & std   & 5.002E-02 & 5.178E-02 & 8.980E-02 & 9.058E-02 & 5.613E-02 & 9.628E-02 \\
    \hline
    \multirow{2}[0]{*}{LIR-CMOP13} & mean  & \textbf{6.455E-02} & 6.460E-02 & 1.200E+00$\dag$ & 1.210E+00$\dag$ & 8.864E-01$\dag$ & 1.180E+00$\dag$ \\
          & std   & 2.177E-03 & 1.639E-03 & 3.055E-01 & 3.172E-01 & 5.756E-01 & 3.775E-01 \\
    \hline
    \multirow{2}[0]{*}{LIR-CMOP14} & mean  & \textbf{6.419E-02} & 6.540E-02$\dag$ & 1.021E+00$\dag$ & 1.107E+00$\dag$ & 1.027E+00$\dag$ & 1.250E+00$\dag$ \\
          & std   & 1.690E-03 & 2.037E-03 & 4.855E-01 & 3.976E-01 & 4.700E-01 & 5.298E-02 \\
    \hline
    \multicolumn{2}{c|}{Wilcoxon-Test (S-D-I)} & --   & 8-5-1  & 14-0-0  & 14-0-0  & 14-0-0  & 14-0-0  \\
    \bottomrule
    \end{tabular}%
  \label{tab:igd}\\
    Wilcoxon’s rank sum test at a 0.05 significance level is performed between PPS-MOEA/D and each of the other five CMOEAs. $\dag$ and $\ddag$ denote that the performance of the corresponding algorithm is significantly worse than or better than that of PPS-MOEA/D, respectively. Where ’S-D-I’ indicates PPS-MOEA/D is superior to, not significantly different from or inferior to the corresponding compared CMOEAs. 

\end{table*}%

\begin{table*}[htbp]
  \centering

  \caption{HV results of PPS-MOEA/D and the other five CMOEAs on LIR-CMOP1-14.}
    \begin{tabular}{c|c|c|c|c|c|c|c}
    \toprule
    \multicolumn{2}{c|}{Test Instances} & PPS-MOEA/D   & MOEA/D-IEpsilon & MOEA/D-Epsilon & MOEA/D-CDP & MOEA/D-SR & C-MOEA/D \\
    \hline
    \multirow{2}[0]{*}{LIR-CMOP1} & mean  & \textbf{1.016E+00} & 1.014E+00$\dag$ & 9.6.0E-01$\dag$ & 7.538E-01$\dag$ & 9.956E-01$\dag$ & 7.414E-01$\dag$ \\
          & std   & 1.580E-03 & 2.433E-03 & 3.280E-02 & 8.947E-02 & 2.915E-02 & 1.222E-01 \\
    \hline
    \multirow{2}[0]{*}{LIR-CMOP2} & mean  & \textbf{1.349E+00} & 1.348E+00$\dag$ & 1.283E+00$\dag$ & 1.061E+00$\dag$$\dag$ & 1.337E+00$\dag$ & 1.070E+00$\dag$ \\
          & std   & 1.010E-03 & 1.318E-03 & 2.880E-02 & 1.078E-01 & 1.466E-02 & 9.099E-02 \\
    \hline
    \multirow{2}[0]{*}{LIR-CMOP3} & mean  & \textbf{8.703E-01} & 8.682E-01$\dag$ & 7.976E-01$\dag$ & 4.856E-01$\dag$ & 5.914E-01$\dag$ & 4.707E-01$\dag$ \\
          & std   & 2.650E-03 & 3.919E-03 & 3.929E-02 & 4.305E-02 & 1.072E-01 & 4.089E-02 \\
    \hline
    \multirow{2}[0]{*}{LIR-CMOP4} & mean  & \textbf{1.093E+00} & \textbf{1.093E+00} & 1.021E+00$\dag$ & 7.349E-01$\dag$ & 8.147E-01$\dag$ & 7.308E-01$\dag$ \\
          & std   & 2.467E-03 & 2.458E-03 & 4.191E-02 & 5.436E-02 & 8.699E-02 & 5.162E-02 \\
    \hline    
    \multirow{2}[0]{*}{LIR-CMOP5} & mean  & \textbf{1.462E+00} & 1.461E+00$\dag$ & 4.297E-02$\dag$ & 1.630E-01$\dag$ & 1.817E-01$\dag$ & 9.721E-02$\dag$ \\
          & std   & 2.919E-04 & 1.332E-03 & 2.353E-01 & 4.426E-01 & 4.389E-01 & 3.699E-01 \\
    \hline
    \multirow{2}[0]{*}{LIR-CMOP6} & mean  & \textbf{1.129E+00} & 9.259E-01$\dag$ & 5.398E-02$\dag$ & 1.880E-01$\dag$ & 3.021E-01$\dag$ & 2.328E-02$\dag$ \\
          & std   & 1.771E-04 & 4.235E-01 & 2.214E-01 & 3.873E-01 & 4.616E-01 & 1.275E-01 \\
    \hline
    \multirow{2}[0]{*}{LIR-CMOP7} & mean  & \textbf{3.015E+00} & 2.863E+00$\dag$ & 3.031E-01$\dag$ & 3.743E-01$\dag$ & 9.878E-01$\dag$ & 2.035E-01$\dag$ \\
          & std   & 2.663E-03 & 1.958E-01 & 9.070E-01 & 9.577E-01 & 1.271E+00 & 7.522E-01 \\
    \hline
    \multirow{2}[0]{*}{LIR-CMOP8} & mean  & \textbf{3.017E+00} & 2.936E+00$\dag$ & 1.060E-01$\dag$ & 5.173E-01$\dag$ & 1.099E+00$\dag$ & 1.658E-01$\dag$ \\
          & std   & 1.139E-03 & 1.855E-01 & 5.488E-01 & 1.054E+00 & 1.198E+00 & 6.114E-01 \\
    \hline
    \multirow{2}[0]{*}{LIR-CMOP9} & mean  & 3.570E+00 & \textbf{3.707E+00}$\ddag$ & 2.737E+00$\dag$ & 2.770E+00$\dag$ & 2.752E+00$\dag$ & 2.770E+00$\dag$ \\
          & std   & 2.242E-01 & 1.878E-02 & 1.483E-01 & 1.843E-01 & 1.640E-01 & 1.841E-01 \\
    \hline
    \multirow{2}[0]{*}{LIR-CMOP10} & mean  & \textbf{3.241E+00} & \textbf{3.241E+00} & 2.885E+00$\dag$ & 2.879E+00$\dag$ & 2.927E+00$\dag$ & 2.888E+00$\dag$ \\
          & std   & 3.077E-04 & 2.482E-04 & 1.023E-01 & 1.364E-01 & 1.347E-01 & 9.765E-02 \\
    \hline
    \multirow{2}[0]{*}{LIR-CMOP11} & mean  & \textbf{4.390E+00} & 4.232E+00 & 3.341E+00$\dag$ & 3.353E+00$\dag$ & 3.383E+00$\dag$ & 3.243E+00$\dag$ \\
          & std   & 2.217E-04 & 1.840E-01 & 2.573E-01 & 2.568E-01 & 2.899E-01 & 2.551E-01 \\
    \hline
    \multirow{2}[0]{*}{LIR-CMOP12} & mean  & \textbf{5.614E+00} & 6.093E+00 & 4.884E+00$\dag$ & 4.827E+00$\dag$ & 5.032E+00 $\dag$ & 4.890E+00 $\dag$ \\
          & std   & 1.6.0E-01 & 1.578E-01 & 3.168E-01 & 3.284E-01 & 1.755E-01 & 3.453E-01 \\
    \hline
    \multirow{2}[0]{*}{LIR-CMOP13} & mean  & 5.710E+00 & \textbf{5.711E+00} & 4.6.0E-01$\dag$ & 4.628E-01$\dag$ & 1.892E+00$\dag$ & 6.293E-01$\dag$ \\
          & std   & 1.275E-02 & 1.301E-02 & 1.301E+00 & 1.423E+00 & 2.572E+00 & 1.712E+00 \\
    \hline
    \multirow{2}[0]{*}{LIR-CMOP14} & mean  & \textbf{6.193E+00} & 6.182E+00$\dag$ & 1.334E+00$\dag$ & 8.813E-01$\dag$ & 1.270E+00$\dag$ & 1.795E-01$\dag$ \\
          & std   & 1.310E-02 & 1.093E-02 & 2.452E+00 & 1.969E+00 & 2.288E+00 & 2.598E-01 \\
    \hline
    \multicolumn{2}{c|}{Wilcoxon-Test (S-D-I)} & --   & 8-5-1  & 14-0-0  & 14-0-0  & 14-0-0  & 14-0-0  \\
    \bottomrule
    \end{tabular}%
  \label{tab:hv}\\
      Wilcoxon’s rank sum test at a 0.05 significance level is performed between PPS-MOEA/D and each of the other five CMOEAs. $\dag$ and $\ddag$ denote that the performance of the corresponding algorithm is significantly worse than or better than that of PPS-MOEA/D, respectively. Where ’S-D-I’ indicates PPS-MOEA/D is superior to, not significantly different from or inferior to the corresponding compared CMOEAs. 
\end{table*}%

\section{Conclusion} \label{sec:conc}
This paper proposes a general PPS framework to deal with CMOPs. More specifically, the search process of PPS is divided into two stages, including the push and pull search processes. At the push stage, constraints are ignored, which can help PPS to get across infeasible regions in front of the unconstrained PFs. Moreover, the landscape affected by constraints can be estimated at the push stage, and this information, such as the ration of feasible solutions and the maximum overall constraint violation, can be applied to conduct the settings of parameters coming from the constraint-handling mechanisms in the pull stage. When the max rate of change between ideal and nadir points is less or equal than a predefined threshold, PPS is switched to the pull search process. The infeasible solutions achieved in the push stage are pulled to the feasible and non-dominated area by adopting an improved epsilon constraint-handling technique. The value of epsilon level can be set properly according to the maximum overall constraint violation obtained at the end of the push search stage. The comprehensive experiments indicate that the proposed PPS achieves competitive or statistically significant better results than the other five CMOEAs on most of the benchmark problems. 

It is also worthwhile to point out that there has been very little work regarding using information of landscape affected by constraints to solve CMOPs. In this context, the proposed PPS provides a viable framework. Obviously, a lot of work need to be done to improve the performance of PPS, such as, the augmented constraint-handling mechanisms in the pull stage, the enhanced strategies to switch the search behavior and the data mining methods and machine learning approaches integrated in the PPS framework. For another future work, the proposed PPS will be implemented in the non-dominated framework, such as NSGA-II, to further verify the effect of PPS. More other CMOPs and real-world constrained engineering optimization problems will also be used to test the performance of the PPS embedded in different MOEA frameworks.


\section*{Acknowledgment}
This research work was supported by Guangdong Key Laboratory of Digital Signal and Image Processing, the National Natural Science Foundation of China under Grant (61175073, 61300159, 61332002, 51375287), Jiangsu Natural Science Foundation (BK20130808) and Science and Technology Planning Project of Guangdong Province, China (2013B011304002).

\ifCLASSOPTIONcaptionsoff
  \newpage
\fi



\bibliographystyle{IEEEtran}

\bibliography{two-stage}
\end{document}